\documentclass{article}

\usepackage{iclr2027_conference,times}
\iclrfinalcopy

\usepackage{amsmath,amssymb,mathtools}
\usepackage{booktabs}
\usepackage{graphicx}
\usepackage{float}
\usepackage{placeins}
\usepackage{xcolor}
\usepackage[hypertexnames=false]{hyperref}
\hypersetup{hidelinks}
\usepackage{url}
\usepackage{multirow}
\usepackage{algorithm}
\usepackage{algpseudocode}
\floatstyle{plain}
\restylefloat{algorithm}
\AddToHook{env/table/begin}{\setlength{\abovecaptionskip}{0pt}\setlength{\belowcaptionskip}{7pt}}

\newcommand{\Lpred}{\mathcal{L}_{\mathrm{pred}}}
\newcommand{\Lcorr}{\mathcal{L}_{\mathrm{corr}}}
\newcommand{\zvec}{\mathbf{z}}
\newcommand{\qvec}{\mathbf{q}}
\newcommand{\sr}[2]{$#1{\scriptstyle\pm}#2$}
\newcommand{\bsr}[2]{{\boldmath$#1{\scriptstyle\pm}#2$}}

\usepackage{amsmath,amssymb}
\usepackage{amsthm}

\theoremstyle{plain}
\newtheorem{theorem}{Theorem}[section]
\newtheorem{proposition}[theorem]{Proposition}

\theoremstyle{definition}

\theoremstyle{remark}
\newtheorem{remark}[theorem]{Remark}

\title{Representation Geometry Matters for\\Planning with JEPA World Models}

\author{\normalfont
  \begingroup
  \setlength{\tabcolsep}{6pt}
  \begin{tabular}{@{}cccc@{}}
    \textbf{Jiaming Hu}\textsuperscript{1,2}\thanks{Correspondence to \texttt{jh7453@bu.edu}. This work was completed while Jiaming Hu was an intern at Unity Technologies.} &
    \textbf{Yan Zheng}\textsuperscript{2} &
    \textbf{Shi Bo}\textsuperscript{2} &
    \textbf{Tian Wang}\textsuperscript{2} \\
    \textbf{Florian Dubost}\textsuperscript{2} &
    \textbf{Alejandro Mottini}\textsuperscript{2} &
    \textbf{Junze Liu}\textsuperscript{2} &
    \textbf{Arvind Srinivasan}\textsuperscript{2} \\
    \textbf{Kai Zhong}\textsuperscript{2} &
    \textbf{Kun Qian}\textsuperscript{2} &
    \textbf{Sharon Gao}\textsuperscript{2} &
    \textbf{Qingjun Cui}\textsuperscript{2} \\[3pt]
    \multicolumn{4}{c}{%
      \textsuperscript{1}Boston University \qquad
      \textsuperscript{2}Unity Technologies} \\[2pt]
    \multicolumn{4}{c}{\scriptsize\texttt{\char123\relax jiaming.hu,yan.zheng,shi.bo,tianwang,}} \\
    \multicolumn{4}{c}{\scriptsize\texttt{florian.dubost,alejandro.mottini,junze.liu,arvind.srinivasan,}} \\[-2pt]
    \multicolumn{4}{c}{\scriptsize\texttt{kai.zhong,kun.qian,sharon.gao,qingjun.cui\char125\relax @unity3d.com}}
  \end{tabular}
  \endgroup
}

\begin{document}

% Use the vertical layout supplied by the ICLR style.

\maketitle
\lhead{Preprint}
\begin{abstract}
Joint-embedding predictive world models support planning through latent predictions, but unconstrained joint training can collapse distinct observations to identical embeddings.
Two prominent strategies for avoiding collapse are to inherit pretrained features, as in DINO-WM, or to learn representations end-to-end with anti-collapse regularization, as in LeWorldModel (LeWM).
Yet avoiding collapse does not ensure that latent distances distinguish outcomes in ways that matter for the task.
In object manipulation, for example, success depends on the object's position and orientation relative to the goal.
Such task-relevant state information can remain accurately decodable while barely influencing latent distance. The resulting planning cost may fail to reflect how close a predicted outcome is to the task goal.
In this paper, we propose SCALE (State-CAlibrated Latent Embeddings), a method that correlates sampled pairwise latent distances with distances in task-relevant state space.
Added to LeWM's existing objective, SCALE preserves its architecture, requires privileged state only during training, and adds no planning-time computation.
We show that SCALE improves planning success over LeWM across manipulation and navigation tasks with multiple solvers and provide a comprehensive analysis of how SCALE reshapes representation geometry to support planning.
\end{abstract}

\section{Introduction}
\label{sec:introduction}

Learning representations that support both prediction and planning is a central challenge in visual control. Joint-embedding predictive architectures (JEPAs) learn by predicting the representation of a target from an observed context, without reconstructing that target in pixel space \citep{assran2023self}. Action-conditioned JEPA world models apply this principle to temporal prediction: an encoder maps observations to a latent space, and a predictor forecasts how those embeddings evolve under candidate actions \citep{zhou2024dino,sobal2025learningrewardfreeofflinedata,maes2026leworldmodel}. In goal-conditioned latent-space planning, solvers such as CEM or iCEM search over action sequences by minimizing the distance between predicted terminal embeddings and a goal embedding. For this objective to be meaningful, latent distances should reflect task-relevant state differences. For example, in pose-sensitive manipulation tasks, observations with similar object poses should have nearby embeddings, whereas larger task-relevant pose differences should correspond to greater latent separation. Candidate scores therefore
depend on how the representation expresses differences between outcomes.  However, accurately decodable task-state information does not necessarily imply such representation geometry: task-relevant differences may contribute little to Euclidean distance and, consequently, to the costs used by the planner.

Existing latent world models obtain their representations in different ways. DINO-WM builds on pretrained visual features, whereas LeWorldModel (LeWM) learns its encoder end-to-end with SIGReg anti-collapse regularization \citep{zhou2024dino,maes2026leworldmodel,balestriero2025lejepa}. Our held-out diagnostics in Sec. \ref{sec:geometry} reveal that DINO-WM exhibits stronger alignment between latent and task-state distances on four of five tasks, even though task-relevant state remains readily decodable from LeWM's representations. This contrast highlights a distinction between preserving state information and making that information influential in the metric used for planning. To characterize this distinction, we introduce \emph{metric leverage}, defined as a latent direction's expected contribution to squared pairwise Euclidean distance. State information can therefore be accurately recovered by a decoder while residing in directions that contribute little to the planning cost. We formalize this separation through a constructive counterexample in Sec. \ref{sec:geometry}, showing that even exact state decodability does not guarantee distance alignment.

\begin{figure}[t]
  \centering
  \includegraphics[width=\linewidth]{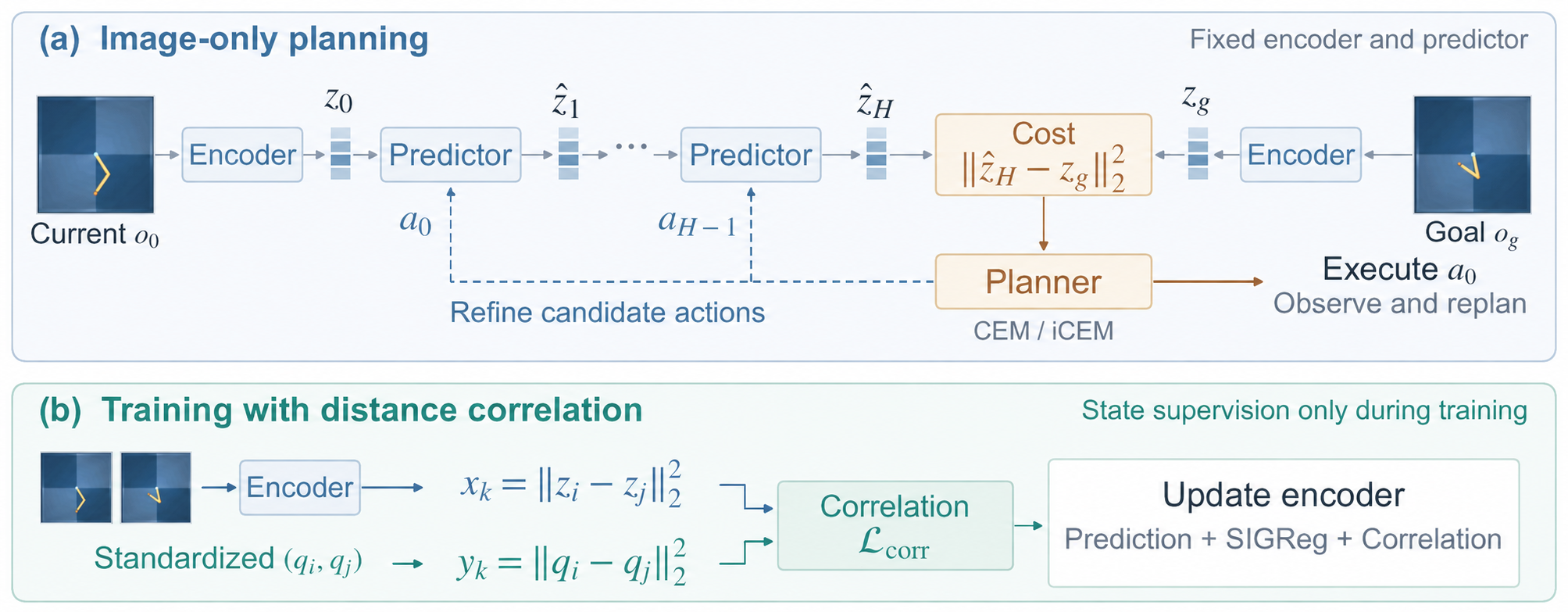}
  \caption{(a) A fixed encoder and predictor support action search using the terminal--goal latent distance. The planner executes the first action block and replans. (b) During training, SCALE correlates sampled latent and task-state distance using the correlation-based regularizer. }
  \label{fig:scale-planning}
  \vspace{-0.5em}
\end{figure}

Motivated by this contrast, we seek to transfer the favorable geometric property observed in DINO-WM without inheriting its computational cost. In our implementation, DINO-WM requires approximately $75\times$ the planning time of LeWM at the largest matched planning budget (Appendix~\ref{app:compute-cost}). We therefore propose SCALE (State-CAlibrated Latent Embeddings), a lightweight training-time method that augments LeWM's existing objective (Fig.~\ref{fig:scale-planning}). SCALE calibrates geometry by correlating sampled pairwise latent distances with distances in standardized task-relevant state space. Selected simulator variables supply supervision only during training; the architecture and image-only planner are unchanged, with no additional planning-time computation.

We evaluate SCALE on Push-T, Reacher, Cube, Two-Room, and PointMaze using several sampling-based solvers, varying the number of candidate action sequences and search iterations to control planning computation. Averaged over these search settings, SCALE improves planning success over LeWM on all five tasks with each solver. To distinguish geometry calibration from improved state recovery, we introduce an auxiliary state-regression control, Aux, using the same supervision. Aux makes state highly decodable but yields less consistent planning gains. Together, these diagnostics show that planning depends on how task-relevant information shapes latent distance, not merely on whether it is preserved. Our main contributions are as follows:
\begin{itemize}
  \item We characterize \textit{metric leverage} and the separation between state decodability and distance alignment through empirical diagnostics and a constructive counterexample (Sec.~\ref{sec:geometry}).
  \item We introduce SCALE, give a conditional candidate-selection bound separating distance misalignment from rollout error, and provide a comprehensive explanation of how SCALE improves planning by reshaping representation geometry (Secs.~\ref{sec:method} and~\ref{sec:experiments}).
  \item We demonstrate that SCALE consistently improves over LeWM across five tasks and two planning solvers, averaged over multiple compute budgets, and use a matched state-regression baseline to investigate the benefits of distance alignment beyond improved state decodability (Sec.~\ref{sec:experiments}).
\end{itemize}

\section{Related Work}
\label{sec:related-work}
\paragraph{Predictive representations and collapse prevention.}
JEPA learns representations by predicting target embeddings from an observed context; I-JEPA applies this principle to masked image regions \citep{assran2023self}. Action-conditioned world models extend the prediction task to future observations. Because the encoder and predictor can trivially agree by mapping every observation to a constant, the representation-learning objective must also prevent collapse. One approach fixes a representation learned beforehand: DINO-WM predicts DINOv2 patch features, inheriting their visual structure while keeping the encoder frozen \citep{oquab2023dinov2,zhou2024dino}. Another learns the representation together with dynamics. PLDM combines latent prediction with VICReg-inspired regularization and inverse dynamics \citep{bardes2021vicreg,sobal2025learningrewardfreeofflinedata}, whereas LeWM uses next-embedding prediction and SIGReg, which encourages an isotropic Gaussian embedding distribution \citep{balestriero2025lejepa,maes2026leworldmodel}. LpWM explores non-negative sparse representations as an alternative to LeWM's dense Gaussian-matched embeddings \citep{kuang2026lpwm}. These designs determine how information is encoded and how collapse is avoided. SCALE retains LeWM's learning framework and adds supervision for a separate property: whether latent distances reflect differences in the selected task state.

\paragraph{Supervision for planning.}
Several approaches add supervision to make JEPA representations
more useful for planning. Value-guided action planning aligns latent distances with goal-conditioned value \citep{destrade2025value}. SCALE instead uses pairwise distances in logged task state as its supervision signal, without requiring reward labels, a learned value function, or a policy. State-grounding methods provide another closely related comparison. PhyLatent diagnoses physically separated states that remain close in LeWM's latent space and incorporates physical-state supervision \citep{zeng2026phylatent}; PSG-JEPA includes auxiliary objectives for physical state and state changes \citep{yan2026forward}. Their state-prediction heads make physical information accessible through a learned readout. Accurate recovery through such a head, however, does not by itself require that information to influence the Euclidean distance used by the planner.  Our Aux control isolates this supervision mechanism. Appendix~\ref{app:extended-related-work} reviews planning-aware world models, behavioral metrics, and plannable representations.

\section{Preliminaries}
\label{sec:preliminaries}

\subsection{Latent-space planning}
\label{sec:planning-objective}

Let $o_t$ be an image frame and $\mathbf a_t$ an action. An episode is one recorded environment trajectory; sub-trajectories are its contiguous segments. LeWM uses a ViT-Tiny encoder $E_\theta$ and an action-conditioned causal Transformer predictor $P_\psi$ (Appendix~\ref{app:impl}). They produce $\zvec_t=E_\theta(o_t)\in\mathbb R^{d_z}$ and $\hat{\zvec}_{t+1}=P_\psi(\zvec_t,\mathbf a_t)$. At test time, $H$ is the planning horizon in model steps. Given candidate actions $\mathbf a_{0:H-1}$ and current frame $o_0$, the model sets $\hat{\zvec}_0=E_\theta(o_0)$ and rolls out
\begin{equation}
\hat{\zvec}_{t+1}=P_\psi(\hat{\zvec}_t,\mathbf{a}_t),
\qquad t=0,\ldots,H-1 .
\end{equation}
We suppress the predictor's finite history in this notation; $t$ indexes model steps and $\mathbf a_t$ denotes an action block. For a fixed initial observation and history, let $\hat z_H(a_{0:H-1})$ denote the terminal embedding predicted by rolling out the model under the candidate action sequence $a_{0:H-1}$. For a goal observation $o_g$ with embedding $\zvec_g=E_\theta(o_g)$, the planner ranks candidates by a cost function
\begin{equation}
J(\mathbf{a}_{0:H-1}) = C\big(\hat{\zvec}_H(\mathbf{a}_{0:H-1}),\, \zvec_g\big) ,
\label{eq:general-cost}
\end{equation}
where $C$ scores a terminal--goal embedding pair. The planner is the optimizer that searches over action sequences using $J$. Following LeWM and DINO-WM, we adopt the squared Euclidean distance as the cost function:
\begin{equation}
J(\mathbf{a}_{0:H-1}) = \big\lVert \hat{\zvec}_H(\mathbf{a}_{0:H-1}) - \zvec_g \big\rVert_2^2 .
\label{eq:planning-cost}
\end{equation}

\subsection{LeWM Training objective}
\label{sec:base}

For a transition $(o_t,\mathbf{a}_t,o_{t+1})$, the encoder and predictor are trained with the one-step latent-prediction loss
\begin{equation}
\Lpred = \big\lVert \hat{\zvec}_{t+1} - \zvec_{t+1} \big\rVert_2^2 ,
\end{equation}
where $\zvec_{t+1}=E_\theta(o_{t+1})$ is the target embedding. This loss alone admits collapsed solutions in which the encoder maps every observation to the same point. For a batch of $N$ encoder outputs $Z=\{\zvec_i\}_{i=1}^{N}$, SIGReg samples random unit directions in latent space, projects $Z$ onto each, and penalizes deviations of the projected empirical distributions from a univariate standard normal \citep{balestriero2025lejepa,maes2026leworldmodel}. Applied across many directions, this drives the embedding distribution toward an isotropic Gaussian and maintains global spread. LeWM therefore optimizes
\begin{equation}
\mathcal{L}_{\mathrm{LeWM}}
= \Lpred + \lambda_{\mathrm{sig}}\,\mathrm{SIGReg}(Z) .
\end{equation}

\section{Where Task-Relevant State Lives}
\label{sec:geometry}

Under the squared Euclidean objective in Eq.~\eqref{eq:planning-cost}, the solver favors candidate action sequences whose predicted terminal embeddings lie closer to the goal embedding. For this comparison to guide task completion, latent distances should reflect differences in task-relevant state. In object manipulation, for example, differences in object position and orientation relative to the target should affect the planning cost. A cost that is insensitive to these differences may provide little guidance toward the task goal. To examine whether the representations capture this relationship, we compute the Spearman rank correlation $\rho_S$ between $\|\mathbf{z}_i-\mathbf{z}_j\|_2^2$ and $\|\mathbf{q}_i-\mathbf{q}_j\|_2^2$ over held-out pairs of image frames, where $\mathbf{q}$ denotes the task-relevant state variables logged alongside each trajectory (formally
specified in Sec. \ref{sec:corr-loss}). We rank the two distance lists separately, assigning average ranks to ties. A high positive correlation indicates that pairs farther apart in task state also tend to be farther apart in latent space.

\begin{table}[H]
  \centering
  \caption{Spearman correlation $\rho_{\mathrm{S}}$ between latent and task-state distances for LeWM and DINO-WM on held-out frame pairs.}
  \label{tab:ref-align}
  \small
  \begin{tabular}{lcc}
    \toprule
    Task & LeWM & DINO-WM \\
    \midrule
    Push-T    & .13 & .70 \\
    Reacher   & .18 & .46 \\
    Cube      & .00 & .46 \\
    Two-Room  & .41 & .85 \\
    PointMaze & .52 & .50 \\
    \bottomrule
  \end{tabular}
  \normalsize
  \setlength{\abovecaptionskip}{7pt}
  \setlength{\belowcaptionskip}{0pt}
\end{table}
Table~\ref{tab:ref-align} shows higher distance alignment for DINO-WM on four tasks and similar alignment on PointMaze. Yet probes (i.e., regression models trained to predict state variables from frozen embeddings) recover most task-relevant state accurately from both full embeddings (see Appendix~\ref{app:representation} for more details). Why can state be recoverable while distances reflect it poorly? Proposition~\ref{prop:metric_decodability} makes this separation precise: even perfect linear recovery need not imply aligned distances. To see why a decoder can recover but a fixed Euclidean cost can underweight, we examine each direction's contribution to that cost.

Let $z\in\mathbb R^{d_z}$ be a square-integrable encoder output, mean $\mu=\mathbb E[z]$, and covariance $\Sigma_z=\mathbb E[(z-\mu)(z-\mu)^\top]=U\Lambda U^\top$. Here $U^\top U=I_{d_z}$, $\Lambda=\operatorname{diag}(\lambda_1,\ldots,\lambda_{d_z})$, and $\lambda_1\ge\cdots\ge\lambda_{d_z}\ge0$. Writing $U=(\mathbf u_1,\ldots,\mathbf u_{d_z})$ and $c=U^\top(z-\mu)$ gives
\begin{equation}
z-\mu=\sum_{j=1}^{d_z}c_j\mathbf u_j,
\qquad \mathrm{Var}(c_j)=\lambda_j.
\end{equation}
For any two embeddings $\zvec_i$ and $\zvec_g$, orthogonality yields the exact decomposition
\begin{equation}
\big\lVert\zvec_i-\zvec_g\big\rVert_2^2
=\sum_{j=1}^{d_z}\big(c_{ij}-c_{gj}\big)^2.
\label{eq:cost-decomposition}
\end{equation}
We define the \textit{metric leverage} of direction $j$ as its expected contribution to squared distance between independent copies $z,z'$ of the encoder output, namely $\mathbb E[(c_j-c_j')^2]$. The following proposition relates this quantity to variance and shows that accurate state recovery alone does not guarantee distance alignment. 

\begin{proposition}[Metric leverage and state decodability]
\label{prop:metric_decodability}
For independent copies $z,z'$, with principal coordinates $c,c'$,
the metric leverage of coordinate $j$ is
\begin{equation}
\mathbb E[(c_j-c_j')^2]=2\lambda_j.
\label{eq:leverage}
\end{equation}
Nevertheless, for any integers $d_z>d_q\ge1$ and any $\delta>0$,
there exists a centered jointly Gaussian pair $(q,z)$ with
$\Sigma_q:=\mathbb E[qq^\top]=I_{d_q}$,
$\Sigma_z\succ0$, and $\mathbb E\|z-z'\|_2^2\ge1$, such that
$h(z)=q$ almost surely for some linear decoder $h$, yet
\[
0\le\rho_{\mathrm{P}}(\|z-z'\|_2^2,\|q-q'\|_2^2)<\delta.
\]
Here $(q',z')$ is an independent copy of $(q,z)$ and $\rho_{\mathrm{P}}$ denotes Pearson correlation, distinct from the Spearman diagnostic in Table~\ref{tab:ref-align}.
\end{proposition}

\begin{remark}
The Gaussianity assumption in Proposition \ref{prop:metric_decodability} is used for expositional convenience; the separation it illustrates is not specific to the Gaussian distribution itself. Proposition \ref{prop:metric_decodability} shows that a decoder can rescale the coordinates encoding state to recover it exactly, even when those coordinates contribute arbitrarily little to the squared latent distance on average. Thus,
\[
\qvec\text{ is decodable from }\zvec
\quad\nRightarrow\quad
\qvec\text{ strongly affects }C(\zvec,\zvec_g)
\text{ on average}.
\]
The proof of Proposition~\ref{prop:metric_decodability} is given in Appendix~\ref{app:proofs}.
\end{remark}

Together, Table~\ref{tab:ref-align} and Proposition~\ref{prop:metric_decodability} motivate explicitly relating latent distances to task-state differences during training, beyond making state decodable. We refer to this as \emph{distance calibration}. SCALE implements this through the correlation regularizer $\Lcorr$ (see more details in Sec. \ref{sec:method}), which encourages sampled pairwise latent distances to correlate with the corresponding task-state distances while retaining LeWM's prediction objective and planning procedure.

\section{Method}
\label{sec:method}
\label{sec:corr-loss}
In this section, we formally present our methodology. Consider a minibatch of $B$ sub-trajectories, each containing $T$ frames. We flatten its $N=BT$ frames into $\{(\zvec_n,\qvec_n,b_n,e_n)\}_{n=1}^{N}$, where $\zvec_n\in\mathbb{R}^{\textcolor{black}{d_z}}$ is an encoder output, $\qvec_n\in\mathbb{R}^{d_q}$ is its standardized task-relevant state, $b_n$ identifies its sub-trajectory, and $e_n$ its source episode. At each step, we \textcolor{black}{sample an indexed list $\mathcal{P}=((i_k,j_k))_{k=1}^{K}$ of non-self frame pairs ($i_k\ne j_k$), allowing repeated draws,} with half drawn within sub-trajectories ($b_{i_k}=b_{j_k}$) and half across episodes ($e_{i_k}\neq e_{j_k}$). This stratification supplies both local and global geometric comparisons. We then define the latent and state distance profiles
\begin{equation}
x_k = \lVert \zvec_{i_k} - \zvec_{j_k} \rVert_2^2, \qquad
y_k = \lVert \qvec_{i_k} - \qvec_{j_k} \rVert_2^2 .
\label{eq:distance-profiles}
\end{equation}
Let $\bar{x}$ and $s_x$ be the mean and standard deviation of $\{x_k\}_{k=1}^{K}$; $\bar{y}$ and $s_y$ are defined analogously. With a numerical constant $\varepsilon>0$, the standardized profiles and correlation loss are
\begin{equation}
\tilde{x}_k=\frac{x_k-\bar{x}}{s_x+\varepsilon},
\qquad
\tilde{y}_k=\frac{y_k-\bar{y}}{s_y+\varepsilon},
\qquad
\Lcorr = 1-\frac{1}{K}\sum_{k=1}^{K}\tilde{x}_k\tilde{y}_k .
\label{eq:corr}
\end{equation}
Component-wise standardization of $\qvec$ places the selected state variables on comparable scales, while standardizing the two distance profiles focuses $\Lcorr$ on relative separation rather than absolute scale. Minimizing $\Lcorr$ promotes positive correlation between the squared latent distances $x_k$ and the corresponding squared task-state distances $y_k$, encouraging pairs with larger task-state differences to lie farther apart in latent space. Appendix~\ref{app:q-selection} details the task-specific choices of $\qvec$ and their full-state comparisons.

\textcolor{black}{Therefore, the complete objective is defined as:}
\begin{equation}
\mathcal{L}_{\mathrm{SCALE}} \;=\; \Lpred \;+\; \lambda_{\mathrm{sig}} \, \mathrm{SIGReg}(Z) \;+\; \lambda_{\mathrm{corr}} \, \Lcorr.
\label{eq:total}
\end{equation}
\textcolor{black}{The prediction term learns action-conditioned latent dynamics, SIGReg maintains global spread, and $\Lcorr$ directly supervises the relation between latent and task-state distances.} The complete $\mathcal{L}_{\mathrm{SCALE}}$ updates the encoder $E_\theta$, whereas the predictor $P_\psi$ is updated only by $\Lpred$. Training pseudocode is provided in Appendix~\ref{app:scale-pseudocode}.

\paragraph{From distance calibration to planning.}
By encouraging latent distances to reflect task-state differences, SCALE aims to make the planning cost more informative about which candidate outcomes are closer to the goal. The following result relates candidate selection to distance correlation and rollout prediction error.

Fix a start--goal pair and ${\color{black}M}\ge2$ candidate action sequences. For candidate $m$, let $z_H^{(m),\mathrm{true}}$ and $q_H^{(m),\mathrm{true}}$ denote its encoded simulator terminal observation and terminal task state, respectively, and let $\hat z_H^{(m)}$ be its predicted terminal embedding. For goal embedding $z_g$ and task state $q_g$, write
\[
\hat J_m=\|\hat z_H^{(m)}-z_g\|_2^2,
\qquad
J_m^{\mathrm{true}}=\|z_H^{(m),\mathrm{true}}-z_g\|_2^2.
\]
Let $s_z,s_q$ denote the empirical standard deviations of $\{J_m^{\mathrm{true}}\}_{m=1}^{\color{black}M}$ and $\{\|q_H^{(m),\mathrm{true}}-q_g\|_2^2\}_{m=1}^{\color{black}M}$, respectively, using divisor ${\color{black}M}$ and $\textcolor{black}{\rho_{\mathrm{P}}}$ be the Pearson correlation between these two lists.

\begin{samepage}
\begin{theorem}[Distance calibration and candidate selection]
\label{thm:scale_candidate_selection}
Suppose $s_z,s_q>0$, and set
$C_z=\max_{1\le m\le {\color{black}M}}\|z_H^{(m),\mathrm{true}}-z_g\|_2$
and
$C_{\mathrm{roll}}
=\max_{1\le m\le {\color{black}M}}
\|\hat z_H^{(m)}-z_H^{(m),\mathrm{true}}\|_2$.
Then every
$\hat m\in\operatorname*{arg\,min}_{1\le m\le {\color{black}M}}\hat J_m$
satisfies
\begin{equation}
\label{eq:scale_selection_bound}
\begin{aligned}
0
&\le
\|q_H^{(\hat m),\mathrm{true}}-q_g\|_2^2
-
\min_{1\le m\le {\color{black}M}}
\|q_H^{(m),\mathrm{true}}-q_g\|_2^2
\\
&\le
2s_q\sqrt{{\color{black}M}(1-\textcolor{black}{\rho_{\mathrm{P}}})}
+
\frac{2s_q}{s_z}
\left(2C_zC_{\mathrm{roll}}+C_{\mathrm{roll}}^2\right).
\end{aligned}
\end{equation}
In particular, if $\textcolor{black}{\rho_{\mathrm{P}}}=1$ and $C_{\mathrm{roll}}=0$,
then $\hat m$ minimizes the terminal task-state distance
among the candidates.
\end{theorem}
\end{samepage}

The first term reflects imperfect correlation between latent and task-state distances to the goal; the second accounts for rollout error. These terms vanish under perfect correlation and exact rollouts, respectively. The bound measures the selected candidate's excess squared task-state distance over the best candidate in the evaluated set. The proof is in Appendix~\ref{app:proofs}.

\paragraph{An auxiliary state-regression control (Aux).}

To investigate whether the benefits of state supervision arise simply from improved state decodability, we construct an auxiliary state-regression control, denoted Aux. It uses the same backbone, state targets, and training protocol as SCALE, replacing $\Lcorr$ with a regression loss through a training-only head $h_\phi$:
\begin{equation}
\mathcal{L}_{\mathrm{aux}} = \big\lVert h_\phi(\zvec_n) - \qvec_n \big\rVert_2^2
\label{eq:aux}
\end{equation}
The complete objective is
$\mathcal{L}_{\mathrm{pred}}+\lambda_{\mathrm{sig}}\operatorname{SIGReg}(Z)
+\lambda_{\mathrm{aux}}\mathcal{L}_{\mathrm{aux}}$.
Aux is intended as a mechanism control rather than a separately optimized planning method. Unlike distance calibration, the learned head can rescale weak latent coordinates to recover state without requiring them to carry substantial \textit{metric leverage}. Aux therefore tests whether making state readily decodable is sufficient to improve planning geometry. Implementation details and hyperparameters are provided in Appendix~\ref{app:impl}.

\section{Experiments}
\label{sec:experiments}

\subsection{Setup}
\label{sec:setup}

We follow the LeWM goal-conditioned planning protocol and evaluate CEM \citep{de2005tutorial} and iCEM \citep{pinneri2021sample}, with DINO-WM as a pretrained-feature reference. Table~\ref{tab:cross-task} reports planning success pooled over two independently trained runs, each evaluated on six held-out sets of 100 episodes. Table~\ref{tab:rollout-probe} evaluates rollout error and candidate-ranking fidelity. Planner budgets, horizons, evaluation construction, and complete tier-level results are detailed in Appendix~\ref{app:impl} and Appendix~\ref{app:full-results}.

\subsection{Planning performance}
\label{sec:results}

\begin{table}[t]
  \centering
  \caption{Pooled iCEM success rates (\%) across tasks and compute tiers. Each entry reports mean $\pm$ SD over $12$ set-level measurements: two independently trained runs, each evaluated on six held-out sets of 100 episodes. Overall pools T1--T5. Bold marks the best method per column within each task. Appendix~\ref{app:full-results} reports both CEM and iCEM results.}
  \label{tab:cross-task}
  \scriptsize
  \setlength{\tabcolsep}{2.5pt}
  \renewcommand{\arraystretch}{0.92}
  \begin{tabular}{llcccccc}
    \toprule
    Task & Method & T1 & T2 & T3 & T4 & T5 & Overall \\
    \midrule
    \multirow{4}{*}{Push-T}
      & LeWM & \sr{86.0}{2.4} & \sr{79.8}{2.1} & \sr{73.4}{5.9} & \sr{55.2}{4.8} & \sr{25.1}{5.1} & \sr{63.9}{3.0} \\
      & SCALE & \sr{89.1}{2.0} & \bsr{86.3}{2.8} & \bsr{79.7}{3.6} & \sr{61.3}{6.2} & \sr{25.2}{5.5} & \sr{68.3}{2.8} \\
      & Aux & \bsr{89.9}{2.0} & \sr{86.2}{2.3} & \sr{79.6}{3.4} & \bsr{62.2}{5.3} & \bsr{27.6}{5.2} & \bsr{69.1}{2.6} \\
      & DINO-WM & \sr{70.0}{4.5} & \sr{65.3}{4.5} & \sr{58.8}{5.1} & \sr{41.5}{8.8} & \sr{23.1}{3.7} & \sr{51.7}{4.3} \\
    \midrule
    \multirow{4}{*}{Reacher}
      & LeWM & \sr{84.9}{2.6} & \sr{84.2}{3.4} & \sr{83.8}{3.4} & \sr{79.0}{6.1} & \sr{62.2}{3.6} & \sr{78.8}{1.7} \\
      & SCALE & \bsr{86.7}{3.6} & \bsr{87.0}{3.1} & \bsr{87.3}{3.2} & \bsr{81.7}{3.0} & \bsr{64.4}{3.0} & \bsr{81.4}{2.0} \\
      & Aux & \sr{85.0}{4.4} & \sr{86.1}{3.1} & \sr{85.8}{3.5} & \sr{80.1}{3.7} & \sr{62.9}{5.1} & \sr{80.0}{2.2} \\
      & DINO-WM & \sr{78.7}{4.9} & \sr{78.2}{4.8} & \sr{80.8}{3.5} & \sr{76.0}{4.0} & \sr{53.0}{5.5} & \sr{73.3}{3.1} \\
    \midrule
    \multirow{4}{*}{Cube}
      & LeWM & \sr{68.9}{4.9} & \sr{67.1}{3.6} & \sr{68.7}{4.1} & \sr{61.9}{3.9} & \sr{54.9}{5.7} & \sr{64.3}{4.0} \\
      & SCALE & \bsr{76.8}{3.7} & \bsr{76.6}{4.0} & \bsr{75.2}{3.0} & \bsr{65.8}{3.3} & \bsr{59.8}{3.9} & \bsr{70.8}{3.1} \\
      & Aux & \sr{73.8}{3.6} & \sr{73.0}{4.7} & \sr{72.5}{3.8} & \sr{64.8}{3.4} & \sr{57.8}{3.8} & \sr{68.4}{3.5} \\
      & DINO-WM & \sr{69.9}{2.5} & \sr{67.4}{1.6} & \sr{66.0}{1.7} & \sr{57.7}{3.8} & \sr{51.4}{3.6} & \sr{62.5}{1.3} \\
    \midrule
    \multirow{4}{*}{Two-Room}
      & LeWM & \sr{91.1}{1.6} & \sr{92.9}{1.7} & \sr{88.5}{2.0} & \sr{77.3}{2.2} & \sr{62.4}{2.8} & \sr{82.5}{1.2} \\
      & SCALE & \sr{99.9}{0.3} & \sr{99.8}{0.6} & \sr{97.7}{1.2} & \sr{90.2}{3.0} & \sr{76.2}{2.7} & \sr{92.7}{1.0} \\
      & Aux & \sr{92.3}{2.1} & \sr{93.2}{1.7} & \sr{89.2}{2.0} & \sr{78.7}{2.5} & \sr{62.9}{3.6} & \sr{83.3}{1.1} \\
      & DINO-WM & \bsr{100.0}{0.0} & \bsr{100.0}{0.0} & \bsr{99.8}{0.5} & \bsr{98.6}{0.8} & \bsr{92.5}{2.1} & \bsr{98.2}{0.5} \\
    \midrule
    \multirow{4}{*}{PointMaze}
      & LeWM & \sr{84.0}{4.1} & \sr{83.6}{4.3} & \sr{82.4}{3.1} & \sr{81.2}{4.6} & \sr{73.2}{4.8} & \sr{80.9}{3.5} \\
      & SCALE & \sr{87.6}{3.3} & \sr{85.8}{4.5} & \sr{84.6}{3.2} & \sr{81.9}{4.7} & \sr{77.0}{4.4} & \sr{83.4}{2.5} \\
      & Aux & \sr{80.8}{5.1} & \sr{79.5}{5.1} & \sr{78.7}{5.0} & \sr{79.4}{3.7} & \sr{75.0}{5.8} & \sr{78.7}{4.3} \\
      & DINO-WM & \bsr{93.4}{2.6} & \bsr{92.2}{2.6} & \bsr{90.8}{3.4} & \bsr{90.3}{2.6} & \bsr{85.2}{2.6} & \bsr{90.4}{2.1} \\
    \bottomrule
  \end{tabular}
  \normalsize
  \setlength{\abovecaptionskip}{7pt}
  \setlength{\belowcaptionskip}{0pt}
\end{table}

SCALE improves LeWM on every pooled task--solver average (10/10). Under iCEM (Table~\ref{tab:cross-task}), its gains are $4.4$, $2.6$, $6.5$, $10.2$, and $2.5$ percentage points on Push-T, Reacher, Cube, Two-Room, and PointMaze, respectively. The corresponding CEM gains are $2.4$, $0.3$, $3.2$, $11.7$, and $1.4$ points. This consistency supports improvements across both search procedures and all five tasks.

The regression control helps distinguish distance calibration from state decodability. Full-embedding probes confirm that Aux successfully encodes the supervised state, even exceeding SCALE's decodability on Push-T and Cube (Appendix~\ref{app:representation}). Aux improves seven of ten pooled task--solver averages, ties LeWM once, and is lower on both PointMaze averages. It trails SCALE on four of five tasks within each solver; Push-T is the exception. Precise pose recovery may favor Aux there: it has stronger position/orientation probes and slightly lower rollout error (Table~\ref{tab:rollout-probe}).

DINO-WM remains strongest on Two-Room and PointMaze. Its pretrained spatial patch features \citep{zhou2024dino} may preserve navigation-relevant detail; it also has lower rollout error and higher candidate-ranking agreement (Table~\ref{tab:rollout-probe}). 

\subsection{How SCALE reorganizes the latent geometry}
\label{sec:validation}

To examine how SCALE changes the representation, we study its variance spectrum, the state information in its leading directions, and the fidelity of predicted candidate rankings.

\begin{figure}[!htb]
  \centering
  \includegraphics[width=0.32\linewidth]{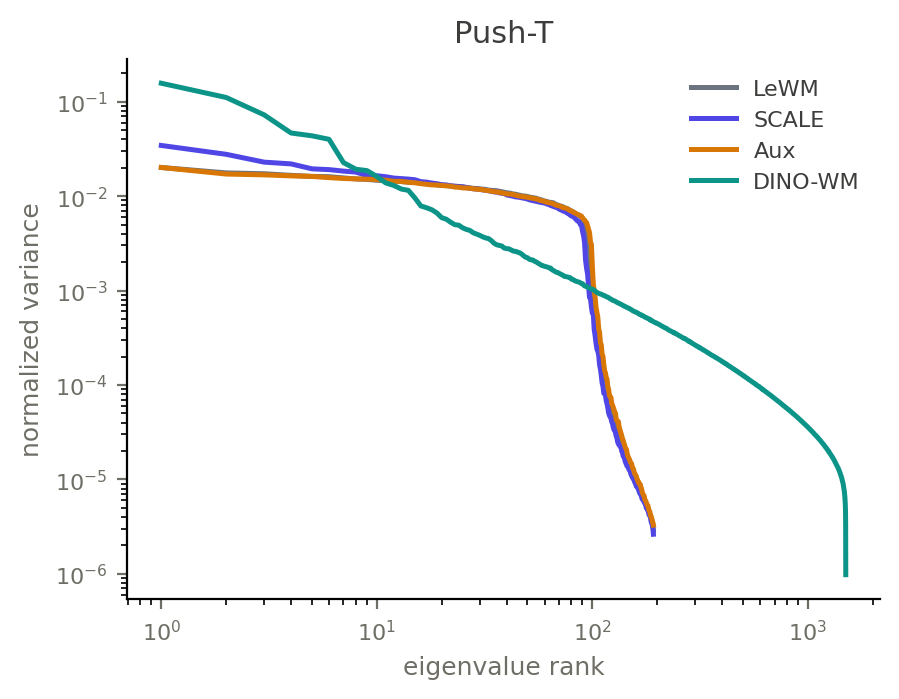}\hfill
  \includegraphics[width=0.32\linewidth]{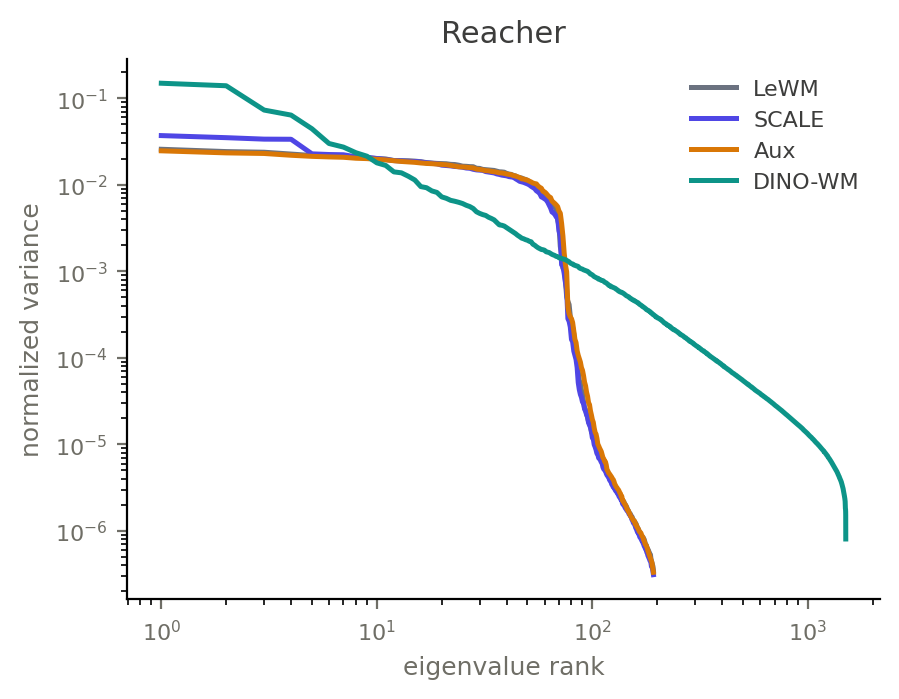}\hfill
  \includegraphics[width=0.32\linewidth]{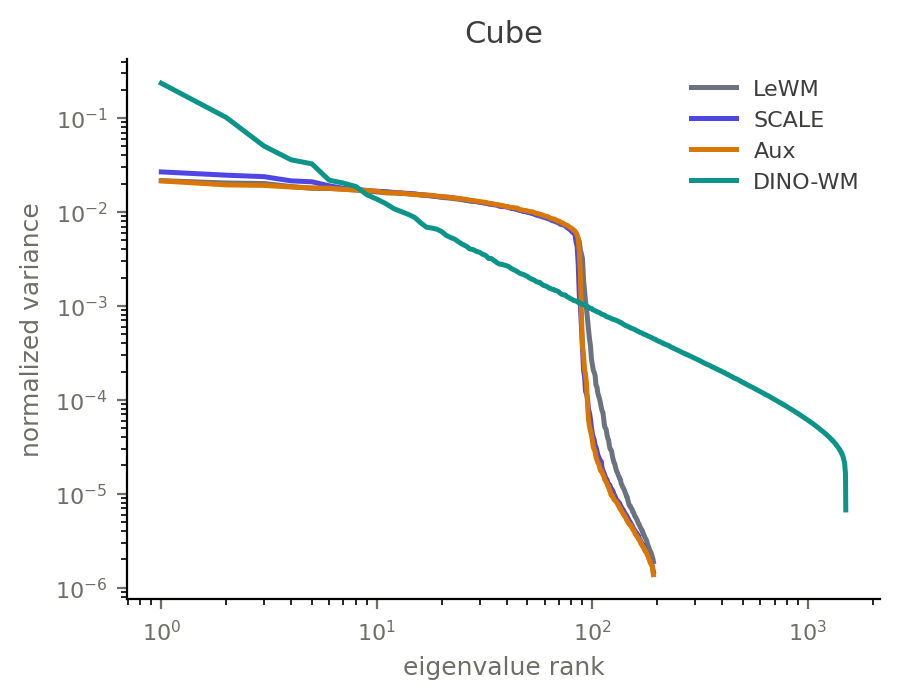}
  \par
  \begin{minipage}[c]{0.32\linewidth}
    \includegraphics[width=\linewidth]{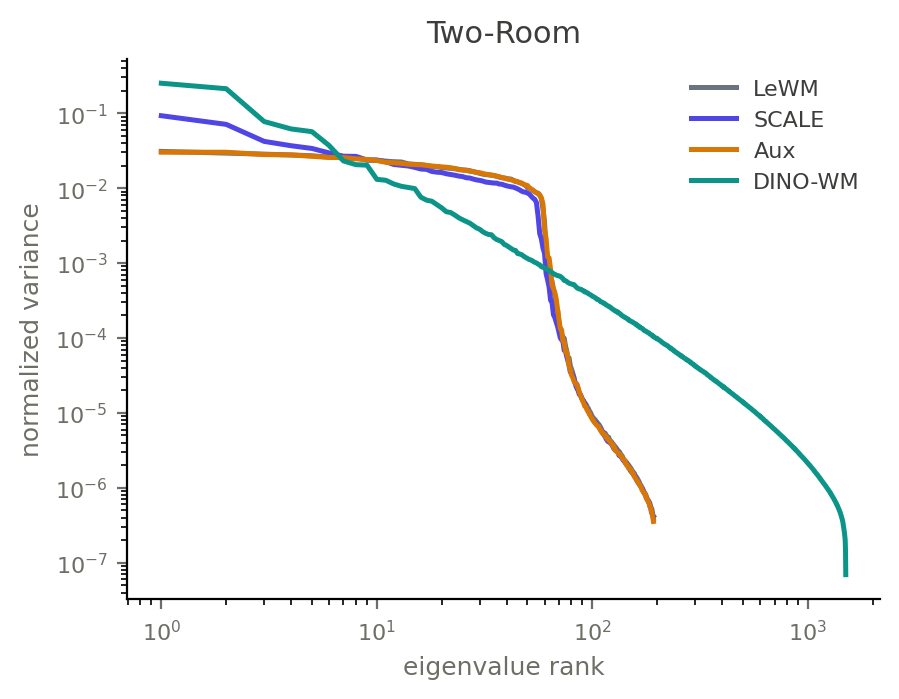}
  \end{minipage}\hfill
  \begin{minipage}[c]{0.32\linewidth}
    \includegraphics[width=\linewidth]{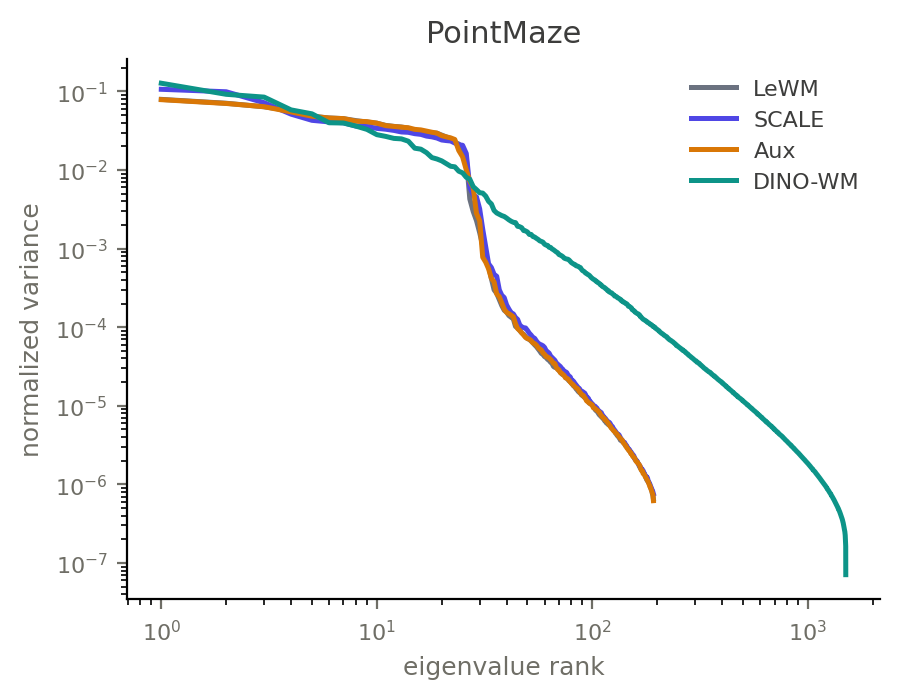}
  \end{minipage}\hfill
  \begin{minipage}[c]{0.29\linewidth}
    \small
    \textcolor[RGB]{107,114,128}{\rule{14pt}{1.2pt}}\quad LeWM\par
    \textcolor[RGB]{79,70,229}{\rule{14pt}{1.2pt}}\quad SCALE\par
    \textcolor[RGB]{217,119,6}{\rule{14pt}{1.2pt}}\quad Aux\par
    \textcolor[RGB]{13,148,136}{\rule{14pt}{1.2pt}}\quad DINO-WM
  \end{minipage}
  \caption{\textcolor{black}{Normalized PCA eigenvalue spectra of LeWM, SCALE, Aux, and DINO-WM across five tasks. Eigenvalues are normalized by total variance, so each principal component's share corresponds to its relative metric leverage. SCALE concentrates more variance in the leading principal components than LeWM and Aux on most tasks.}}
  \label{fig:pca-spectra}
  \vspace{-0.5em}
\end{figure}

\paragraph{SCALE concentrates variance in the leading latent directions.}
Figure~\ref{fig:pca-spectra} shows that SCALE allocates more normalized variance to the leading principal directions than LeWM and Aux on most tasks. DINO-WM also has a strongly weighted spectral head, but its larger representation produces a longer tail. By Sec.~\ref{sec:geometry}, direction $j$ contributes $2\lambda_j$ in expectation to squared distance between independent embeddings. Hence $\lambda_j/\sum_\ell\lambda_\ell$ measures its share of that distance, not absolute latent scale. The spectrum alone does not identify which state variables acquire this leverage.

\begin{figure}[!htb]
  \centering
  \includegraphics[width=0.29\linewidth]{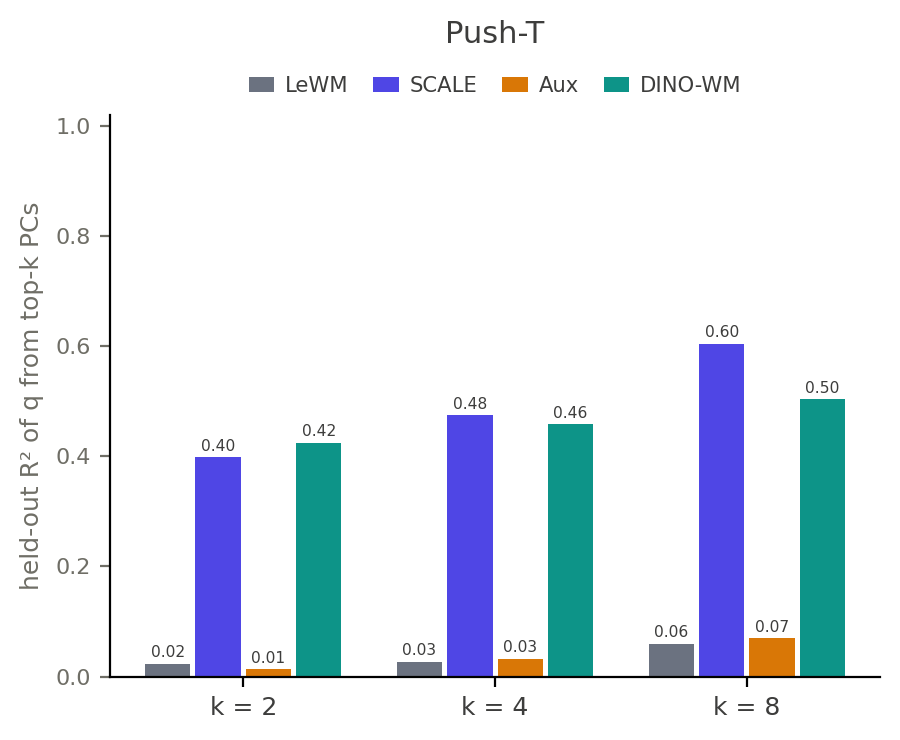}\hfill
  \includegraphics[width=0.29\linewidth]{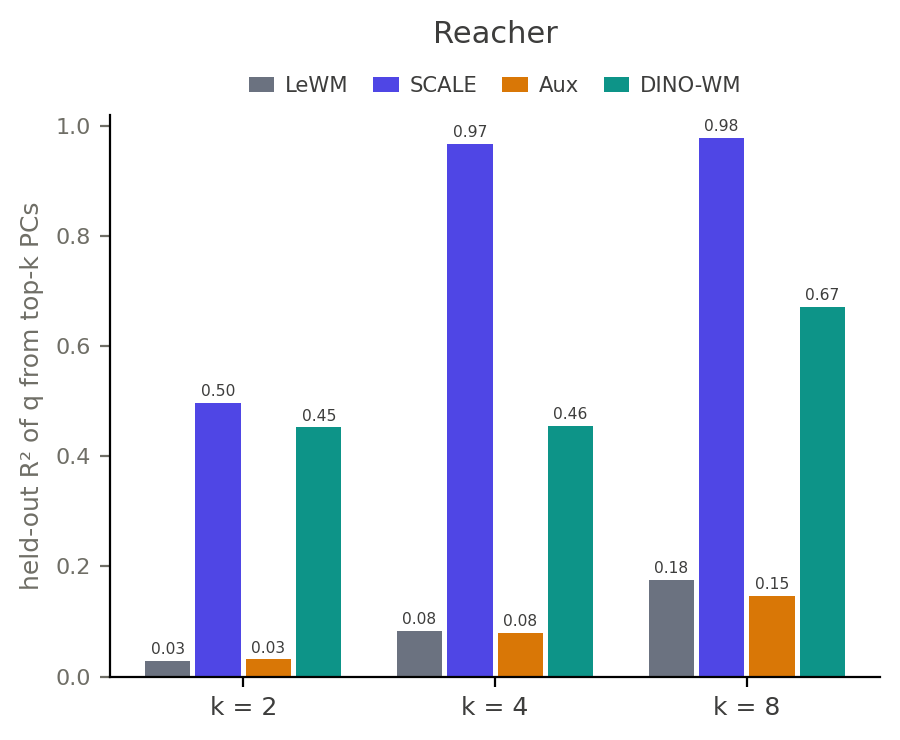}\hfill
  \includegraphics[width=0.29\linewidth]{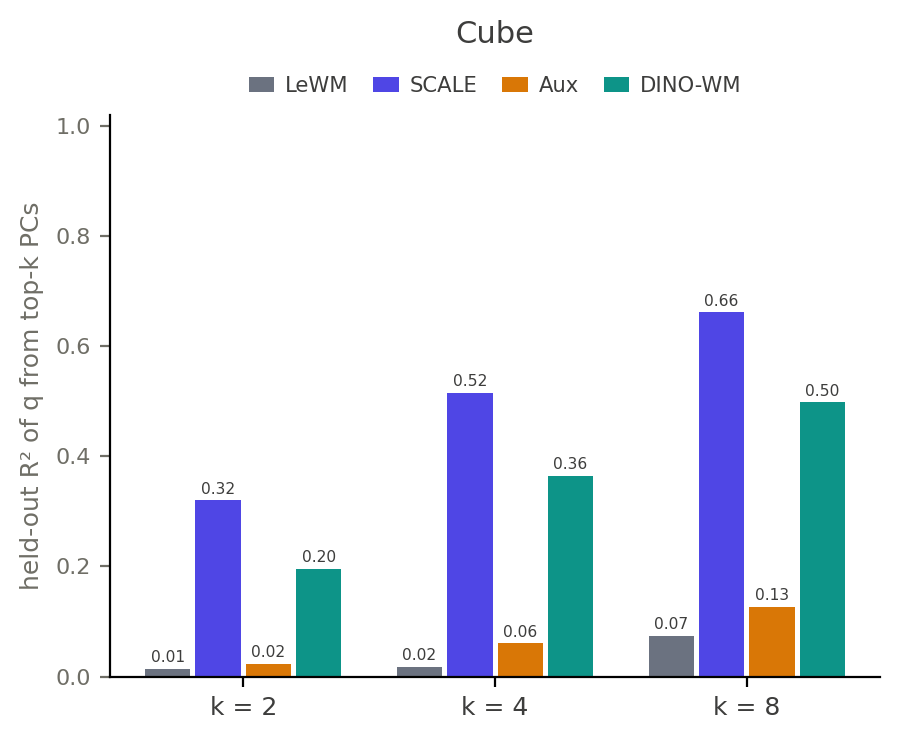}
  \par
  \includegraphics[width=0.29\linewidth]{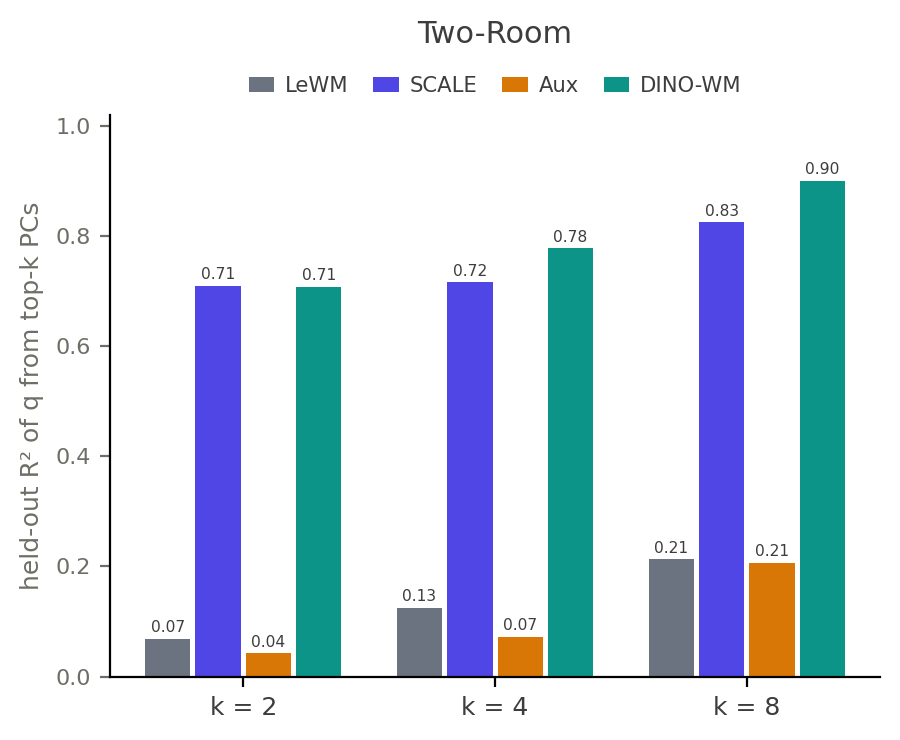}\qquad
  \includegraphics[width=0.29\linewidth]{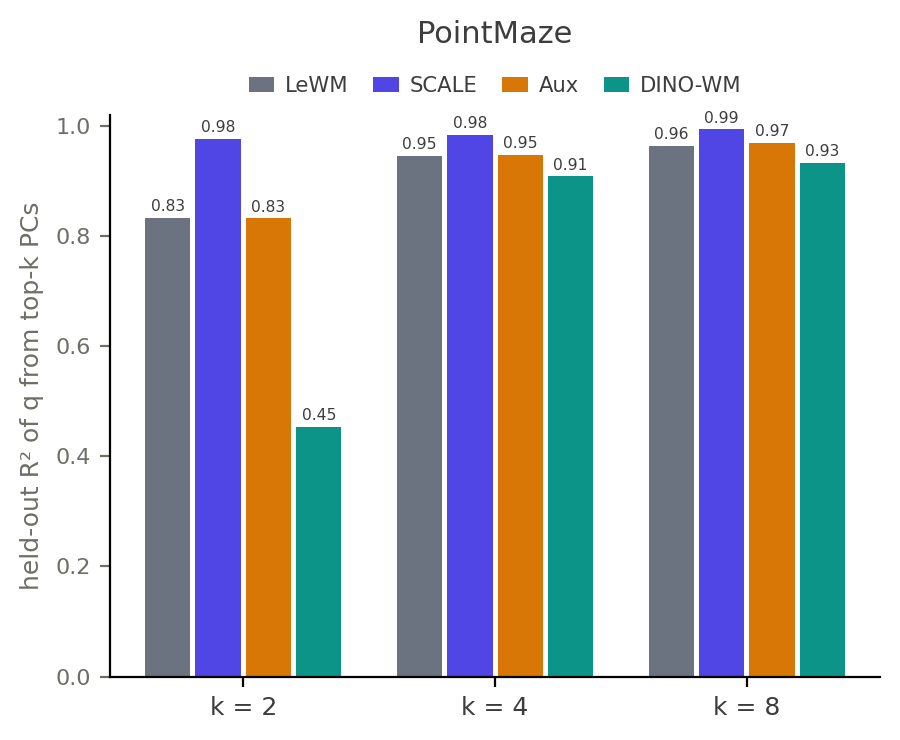}
  \caption{\textcolor{black}{Task-state information in leading principal components. Held-out $R^2$ for predicting task-relevant state from the leading k principal components across five tasks. SCALE improves state predictability compared with LeWM and outperforms Aux on four of five tasks. This suggests that SCALE concentrates task-relevant information in directions with greater metric leverage.}}
  \label{fig:topk-probing}
  \vspace{-0.5em}
\end{figure}

\paragraph{The leading directions encode task-relevant variation.}
We next ask what the high-variance directions actually encode. For each frozen representation, we regress the selected task-relevant state $\qvec$ using only its leading $k$ principal components. Figure~\ref{fig:topk-probing} shows that the leading components of SCALE are substantially more predictive of $\qvec$ than those of LeWM. DINO-WM also concentrates task-state information in leading components: it is comparable to SCALE on Push-T and stronger on Two-Room, but trails SCALE on Reacher, Cube, and PointMaze. The separation from Aux is especially informative: although Aux can recover $\qvec$ accurately from the full embedding, less of that information appears in its leading components on four of the five tasks.

Together, these diagnostics associate the greater concentration of variance in leading directions with more accessible task-state information. Conversely, auxiliary regression can make the same state highly decodable without organizing it in the part of the representation that contributes most strongly to Euclidean distance. Consistent with this interpretation, SCALE also substantially increases held-out latent--state distance alignment, whereas Aux leaves it near the LeWM level despite strong state decodability (Appendix~\ref{app:distance-alignment}).

\paragraph{Rollouts better preserve the encoded candidate ranking.}
The preceding analysis concerns encoded observations, whereas planning operates on predicted future embeddings. For each start and candidate action sequence $m$, the world model predicts a terminal embedding $\hat{\zvec}_H^{(m)}$. Executing the same sequence from the same start in the simulator produces $o_H^{(m),\mathrm{true}}$, which the same encoder maps to $\zvec_H^{(m),\mathrm{true}}=E_\theta(o_H^{(m),\mathrm{true}})$.

These paired outcomes define two measurements. First, rollout error compares the predicted and simulator-derived embeddings as $\lVert\hat{\zvec}_H^{(m)}-\zvec_H^{(m),\mathrm{true}}\rVert_2^2/s$, where $s$ is the model's mean squared distance between random pairs of the simulator-derived terminal embeddings $\zvec_H^{(m),\mathrm{true}}$. Second, for the same goal embedding $\zvec_g$, we compute
\begin{samepage}
\begin{equation*}
\hat{J}_m=\lVert\hat{\zvec}_H^{(m)}-\zvec_g\rVert_2^2,
\qquad
J_m^{\mathrm{true}}=\lVert\zvec_H^{(m),\mathrm{true}}-\zvec_g\rVert_2^2,
\end{equation*}
and report Kendall's $\tau$ between the candidate rankings induced by $\{\hat{J}_m\}$ and $\{J_m^{\mathrm{true}}\}$. We use Kendall's $\tau$ to assess pairwise action-ranking reversals. Spearman's $\rho_{\mathrm S}$ in Table~\ref{tab:ref-align} instead measures monotonic latent--state distance association over observation pairs; the diagnostics answer different questions.
\end{samepage}

\begin{table}[H]
  \centering
  \caption{Within-model rollout quality. Lower rollout error and higher Kendall $\tau$ are better. Parentheses report change relative to LeWM; when LeWM's $\tau$ is near zero, we report the absolute difference $\Delta$. Results average 20 starts.}
  \label{tab:rollout-probe}
  \scriptsize
  \setlength{\tabcolsep}{3pt}
  \resizebox{\linewidth}{!}{%
  \begin{tabular}{llcccc}
    \toprule
    Task & Metric & LeWM & SCALE & Aux & DINO-WM \\
    \midrule
    \multirow{2}{*}{Push-T}
      & RollErr $\downarrow$ & .0350 & .0277 ($-21.0\%$) & \textbf{.0269 ($-23.1\%$)} & .2819 ($+704.9\%$) \\
      & Kendall $\tau$ $\uparrow$ & .7316 & \textbf{.7857 ($+7.4\%$)} & .7576 ($+3.6\%$) & .7000 ($-4.3\%$) \\
    \midrule
    \multirow{2}{*}{Reacher}
      & RollErr $\downarrow$ & .1812 & \textbf{.1548 ($-14.6\%$)} & .1686 ($-7.0\%$) & .5860 ($+223.4\%$) \\
      & Kendall $\tau$ $\uparrow$ & .6233 & \textbf{.6435 ($+3.2\%$)} & .6328 ($+1.5\%$) & .6414 ($+2.9\%$) \\
    \midrule
    \multirow{2}{*}{Cube}
      & RollErr $\downarrow$ & .3195 & .3174 ($-0.7\%$) & \textbf{.2389 ($-25.2\%$)} & .3227 ($+1.0\%$) \\
      & Kendall $\tau$ $\uparrow$ & .3654 & .4433 ($+21.3\%$) & .4150 ($+13.6\%$) & \textbf{.7775 ($+112.8\%$)} \\
    \midrule
    \multirow{2}{*}{Two-Room}
      & RollErr $\downarrow$ & .3600 & .2857 ($-20.6\%$) & .3511 ($-2.5\%$) & \textbf{.1479 ($-58.9\%$)} \\
      & Kendall $\tau$ $\uparrow$ & .5448 & .6210 ($+14.0\%$) & .5389 ($-1.1\%$) & \textbf{.7250 ($+33.1\%$)} \\
    \midrule
    \multirow{2}{*}{PointMaze}
      & RollErr $\downarrow$ & 2.4226 & 2.1132 ($-12.8\%$) & 2.2418 ($-7.5\%$) & \textbf{1.9545 ($-19.3\%$)} \\
      & Kendall $\tau$ $\uparrow$ & $-.0029$ & .1675 (n/a, $\Delta{+}.170$) & .0540 (n/a, $\Delta{+}.057$) & \textbf{.2850 (n/a, $\Delta{+}.288$)} \\
    \bottomrule
  \end{tabular}%
  }
  \normalsize
  \setlength{\abovecaptionskip}{7pt}
  \setlength{\belowcaptionskip}{0pt}
\end{table}

SCALE increases Kendall ranking agreement over LeWM on every task. The improvement is modest on Push-T and Reacher, where LeWM already preserves candidate ordering relatively well, but becomes much larger on Cube, Two-Room, and PointMaze. The higher agreement indicates that rollout predictions better preserve the candidate ordering induced by each model's encoded terminal states. This diagnostic measures fidelity to the model's own latent cost, rather than agreement with task-state distances.

Aux has the lowest rollout error on Push-T and Cube, yet SCALE improves Kendall agreement more on every task. DINO-WM combines favorable rollout and ranking behavior on the navigation tasks, but has much larger rollout errors on Push-T and Reacher; on Cube, it has the highest Kendall $\tau$ without the strongest planning performance. These comparisons distinguish fidelity to a model's latent cost from whether that cost reflects task-state differences, the separate requirements assessed by our rollout and distance-alignment diagnostics.

\section{Conclusion}
\label{sec:conclusion}

Task-relevant state can be decodable from a representation yet weakly influence latent-space planning. Motivated by the contrasting geometries of DINO-WM and LeWM, we introduce SCALE to directly calibrate latent distances with task-relevant state differences during training. Using privileged state only during training, SCALE retains LeWM's end-to-end learning framework and image-only planner while making task-relevant variation more prominent in the latent metric. Planning thus depends on both recoverable information and its influence on the latent cost. Our current $\qvec$ selection follows a simple human prior: use a compact set of static variables while removing velocities and highly redundant coordinates. More broadly, understanding and designing representation geometry for JEPA world models remains an important direction for future research.

\raggedbottom

\section*{AI Use Statement}
Generative AI tools were used to assist with editing, LaTeX formatting, table construction, and the organization, summarization, and presentation of experimental results. 

\section*{Ethics Statement}
This work studies representation learning and planning in simulated benchmark environments. It does not involve human participants, personally identifiable information, or sensitive real-world data. The reported results should therefore not be interpreted as evidence of readiness for safety-critical deployment.

\section*{Reproducibility Statement}
The mathematical formulation of SCALE and its training objective are given in Sec.~\ref{sec:method}, with pseudocode in Appendix~\ref{app:scale-pseudocode}. Appendix~\ref{app:impl} specifies the implementation details. We will publicly release our code to support reproduction and further research.

\bibliography{references}
\bibliographystyle{iclr2027_conference}

\appendix
\raggedbottom

\section{Extended related work}
\label{app:extended-related-work}

\paragraph{Planning-aware latent world models.}
Several recent methods directly address the gap between predictive training and planning-time geometry. Temporal Straightening penalizes local curvature so that Euclidean distance better approximates trajectory geodesics and improves the conditioning of latent planning \citep{wang2026temporal}. RC-aux combines multi-horizon prediction with budget-conditioned reachability supervision and temporal hard negatives, and can expose a reachability-aware planning signal \citep{li2026predictive}. Temporal-Distance JEPA mines directed temporal progress from offline trajectories and uses it both as a learned planning cost and as representation supervision \citep{bai2026temporal}. Trajectory Reachability Metrics instead repair a fixed world model post hoc by learning a horizon-matched pairwise terminal-ranking head from logged trajectory structure \citep{li2026beyond}. Monotone Planning Costs train candidate costs to respect an ordering induced by increasingly perturbed action sequences in image-goal navigation \citep{chahe2026latent}. VIScore takes a complementary diagnostic view, measuring planning-relevant properties across the encoder, predictor, and planner \citep{wu2026viscore}. Complementary work on geometrically regularized world models uses temporal contrastive constraints to improve long-horizon predictive fidelity in deterministic environments, further demonstrating that dynamics accuracy depends on latent geometry \citep{xia2026cloning}. SCALE differs in retaining the original Euclidean image-goal planning interface: it does not deploy a learned cost or reachability head, but calibrates the representation during training so that the existing terminal--goal distance better reflects selected task-state differences.

\paragraph{Behavioral metrics and plannable representations.}
Earlier methods explicitly shape representations using decision-relevant structure. DBC matches latent distances to bisimulation distances, while MICo develops a sample-based behavioral metric \citep{zhang2020learning,castro2021mico}. Plan2vec distills long-range distances from planned paths, and PcLast organizes latent states by reachability \citep{yang2020plan2vec,koul2023pclast}. These methods establish a broader connection between representation geometry and decision making; SCALE applies this principle to the Euclidean image-goal cost used by JEPA world models.

\FloatBarrier
\section{\texorpdfstring{\textcolor{black}{Proof}}{Proof}}
\label{app:proofs}

\begin{proof}[Proof of Proposition~\ref{prop:metric_decodability}]
Since $\mathbb E[c]=0$ and
$\mathbb E[cc^\top]=U^\top\Sigma_z U=\Lambda$, independence gives
\[
\mathbb E[(c_j-c_j')^2]
=
\mathbb E[c_j^2]+\mathbb E[(c_j')^2]
-2\mathbb E[c_j]\mathbb E[c_j']
=
2\lambda_j.
\]

For the counterexample, choose $0<\alpha<1$ and take
\[
z=c\sim N(0,\Sigma_z),
\qquad
\Sigma_z=
\begin{pmatrix}
I_{d_z-d_q} & 0\\
0 & \alpha^2 I_{d_q}
\end{pmatrix}.
\]
Here $U=I_{d_z}$. Define
$q=h(z)=\alpha^{-1}(z_{d_z-d_q+1},\ldots,z_{d_z})^\top$.
Then $h$ is linear, $(q,z)$ is jointly Gaussian,
$\Sigma_q=I_{d_q}$, and $\Sigma_z\succ0$.
Moreover,
$\mathbb E\|z-z'\|_2^2
=2(d_z-d_q+\alpha^2d_q)\ge2$.

The squared latent distance decomposes as
\[
\|z-z'\|_2^2
=
\sum_{j=1}^{d_z-d_q}(c_j-c_j')^2
+
\alpha^2\|q-q'\|_2^2.
\]
{\color{black}
Write the first term as $A$ and set $B=\|q-q'\|_2^2$.
The two Gaussian coordinate blocks are independent, so $A$ and $B$
are independent, with $A/2\sim\chi^2_{d_z-d_q}$ and
$B/2\sim\chi^2_{d_q}$. Hence
$\operatorname{Var}(A)=8(d_z-d_q)$ and
$\operatorname{Var}(B)=8d_q$. In particular,
\begin{align*}
\operatorname{Cov}(A+\alpha^2 B,B)&=8\alpha^2d_q,\\
\operatorname{Var}(A+\alpha^2 B)&=8(d_z-d_q)+8\alpha^4d_q.
\end{align*}
}Consequently,
\begin{align*}
\textcolor{black}{\rho_{\mathrm{P}}}(\|z-z'\|_2^2,\|q-q'\|_2^2)
&=
\frac{8\alpha^2d_q}
{\sqrt{8(d_z-d_q+\alpha^4d_q)\cdot8d_q}}
\\
&=
\left(
1+\frac{d_z-d_q}{\alpha^4d_q}
\right)^{-1/2}
\longrightarrow0
\qquad\text{as }\alpha\downarrow0.
\end{align*}
Choosing $\alpha > 0$ sufficiently small makes
the correlation smaller than $\delta$
while preserving exact linear recovery $h(z) = q$ and the stated nondegeneracy conditions.
In this construction, each task-state coordinate has metric
leverage $2\alpha^2$, despite exact state recovery.
\end{proof}

\begin{proof}[Proof of Theorem~\ref{thm:scale_candidate_selection}]
Choose any
$m^\star\in\operatorname*{arg\,min}_{1\le m\le {\color{black}M}}
\|q_H^{(m),\mathrm{true}}-q_g\|_2^2$.
Expanding the squared norms gives, for every $m$,
\begin{align*}
|\hat J_m-J_m^{\mathrm{true}}|
&\le
2\|z_H^{(m),\mathrm{true}}-z_g\|_2
  \|\hat z_H^{(m)}-z_H^{(m),\mathrm{true}}\|_2
\\
&\quad+
\|\hat z_H^{(m)}-z_H^{(m),\mathrm{true}}\|_2^2
\\
&\le
2C_zC_{\mathrm{roll}}+C_{\mathrm{roll}}^2.
\end{align*}
Since $\hat J_{\hat m}\le\hat J_{m^\star}$, it follows that
\begin{align*}
J_{\hat m}^{\mathrm{true}}-J_{m^\star}^{\mathrm{true}}
&\le
|J_{\hat m}^{\mathrm{true}}-\hat J_{\hat m}|
+
|\hat J_{m^\star}-J_{m^\star}^{\mathrm{true}}|
\\
&\le
2\left(2C_zC_{\mathrm{roll}}+C_{\mathrm{roll}}^2\right).
\end{align*}

Next, write $Q_m=\|q_H^{(m),\mathrm{true}}-q_g\|_2^2$ and define
the standardized vectors $u,v\in\mathbb R^{\color{black}M}$ by
\[
u_m=\frac{J_m^{\mathrm{true}}-\bar J}{s_z},
\qquad
v_m=\frac{Q_m-\bar Q}{s_q},
\]
where $\bar J={\color{black}M}^{-1}\sum_m J_m^{\mathrm{true}}$ and
$\bar Q={\color{black}M}^{-1}\sum_m Q_m$. Because $s_z,s_q$ use divisor ${\color{black}M}$,
$\|u\|_2^2=\|v\|_2^2={\color{black}M}$ and $u^\top v={\color{black}M}\textcolor{black}{\rho_{\mathrm{P}}}$.
Thus $w=v-u$ satisfies $\|w\|_2^2=2{\color{black}M}(1-\textcolor{black}{\rho_{\mathrm{P}}})$.
For standard basis vectors $e_1,\ldots,e_{\color{black}M}$, Cauchy--Schwarz gives
\[
|w_{\hat m}-w_{m^\star}|
=|(e_{\hat m}-e_{m^\star})^\top w|
\le\sqrt{2}\,\|w\|_2
=2\sqrt{{\color{black}M}(1-\textcolor{black}{\rho_{\mathrm{P}}})}.
\]
This inequality also holds when $\hat m=m^\star$.
Centering cancels in the identity
$(Q_{\hat m}-Q_{m^\star})/s_q
=(J_{\hat m}^{\mathrm{true}}-J_{m^\star}^{\mathrm{true}})/s_z
+w_{\hat m}-w_{m^\star}$, so
\begin{align*}
&\frac{
\|q_H^{(\hat m),\mathrm{true}}-q_g\|_2^2
-
\|q_H^{(m^\star),\mathrm{true}}-q_g\|_2^2
}{s_q}
\\
&\qquad\le
\frac{J_{\hat m}^{\mathrm{true}}-J_{m^\star}^{\mathrm{true}}}{s_z}
+
2\sqrt{{\color{black}M}(1-\textcolor{black}{\rho_{\mathrm{P}}})}.
\end{align*}
Multiplying by $s_q$ and applying the preceding latent-cost
bound proves the upper bound in \eqref{eq:scale_selection_bound}.
The lower bound follows from the definition of $m^\star$.
When $\textcolor{black}{\rho_{\mathrm{P}}}=1$ and $C_{\mathrm{roll}}=0$, the upper bound
vanishes, proving the final claim.
\end{proof}

\section{Implementation details}
\label{app:impl}

\subsection{SCALE Pseudocode}
\label{app:scale-pseudocode}

\begingroup
\floatstyle{ruled}
\restylefloat{algorithm}

\begin{algorithm}[htbp]
  \caption{SCALE training.}
  \label{alg:scale}
  \small

  \begin{algorithmic}[1]
    \Require $\mathcal{B}=(O,A,Q,e)$: aligned pixels $O$ and states $Q$
    over $B\times T$ frames, actions $A$ over $B\times(T-1)$ transitions,
    and episode IDs $e=(e_b)_{b=1}^{B}$

    \Require $E_\theta$, $P_\psi$, $K\in2\mathbb{N}$,
    $\varepsilon,\delta_{\mathrm{std}}>0$, and
    $\lambda_{\mathrm{sig}},\lambda_{\mathrm{corr}}\geq0$

    \State $Z\gets E_\theta(O)$;
    $\widehat{Z}\gets P_\psi(Z_{:,1:T-1},A)$

    \State $\Lpred\gets
    \operatorname{MSE}(\widehat{Z},Z_{:,2:T})$

    \State $z\gets\operatorname{reshape}(Z,(N,d_z))$;
    $q\gets\operatorname{detach}\!\left(
      \operatorname{reshape}(Q,(N,d_q))
    \right)$
    \Comment{$N=BT$ frames}

    \State $b_n\gets\lceil n/T\rceil$;
    $e_n\gets e_{b_n}$ for $n=1,\ldots,N$

    \State Sample $K/2$ pairs within sub-trajectories ($b_i=b_j$)
    and $K/2$ across episodes ($e_i\neq e_j$),
    with $i\neq j$ and repeated draws allowed;
    concatenate the draws into $\mathcal{P}$

    \State $x_k\gets\lVert z_{i_k}-z_{j_k}\rVert_2^2$;
    $y_k\gets\lVert q_{i_k}-q_{j_k}\rVert_2^2$
    for $k=1,\ldots,K$

    \If{$\operatorname{std}(x)<\delta_{\mathrm{std}}$
        \textbf{or}
        $\operatorname{std}(y)<\delta_{\mathrm{std}}$}
      \State $\Lcorr\gets0\cdot\operatorname{sum}(z)$
      \Comment{graph-connected zero}
    \Else
      \State $\tilde{x}\gets
      (x-\operatorname{mean}(x))/(\operatorname{std}(x)+\varepsilon)$;
      $\tilde{y}\gets
      (y-\operatorname{mean}(y))/(\operatorname{std}(y)+\varepsilon)$

      \State $\Lcorr\gets
      1-\operatorname{mean}(\tilde{x}\odot\tilde{y})$
    \EndIf

    \State $\mathcal{L}_{\mathrm{SCALE}}\gets
    \Lpred
    +\lambda_{\mathrm{sig}}\operatorname{SIGReg}(Z)
    +\lambda_{\mathrm{corr}}\Lcorr$

    \State $g_\theta\gets\nabla_\theta\mathcal{L}_{\mathrm{SCALE}}$;
    $g_\psi\gets\nabla_\psi\Lpred$

    \State $(\theta,\psi)\gets
    \operatorname{Adam}\bigl((\theta,\psi),(g_\theta,g_\psi)\bigr)$
  \end{algorithmic}
\end{algorithm}

\endgroup

\paragraph{Hyperparameters.} The encoder is a ViT-Tiny with patch size 14, 12 layers, 3 attention heads, and width 192. Its CLS token is passed through a linear--BatchNorm projector. The predictor is a six-layer causal transformer with 16 attention heads, dropout 0.1, a history length of 3, and actions injected through zero-initialized AdaLN; its output uses a matching projector. We use frame skip 5 and concatenate the five intervening actions into action blocks, whose dimension is five times the environment action dimension. Training uses four-frame sub-trajectories, batch size 128, and 10 epochs. We optimize with Adam using learning rate $5\times10^{-5}$ and weight decay $10^{-3}$. SIGReg uses $1024$ random unit projections with $\lambda_{\mathrm{sig}}=0.09$. For $\Lcorr$, we sample $K{=}4096$ pairs per training step with equal within- and cross-episode proportions and set $\varepsilon=10^{-6}$. We use $\lambda_{\mathrm{corr}}=0.1$ for all tasks. For Aux, $h_\phi$ is a two-layer MLP, $192\!\rightarrow\!256\!\rightarrow\!d_q$, with a ReLU after the first linear layer and no normalization; $d_q$ matches the task-state dimension defined in Appendix~\ref{app:q-selection}. We set $\lambda_{\mathrm{aux}}$ to $0.3$, $0.4$, $0.1$, $0.1$, and $0.1$ for Push-T, Reacher, Cube, Two-Room, and PointMaze, respectively. This configuration was selected through limited preliminary checks for high state-probe accuracy, without a large-scale architecture or hyperparameter sweep. All reported SCALE and Aux results use these settings; Appendix~\ref{app:q-selection} evaluates the corresponding full-state alternatives for SCALE.

\paragraph{Planner.} All planning solvers use a horizon of five action blocks (25 environment steps), an action-block size of five, a receding-horizon interval of five, a 50-step evaluation horizon, and goals sampled 25 steps ahead. Compute tiers T1--T5 specify candidate count / optimization iterations as $300/30$, $100/20$, $50/10$, $20/5$, and $10/3$, corresponding to $9000$, $2000$, $500$, $100$, and $30$ rollout evaluations per replanning step. The elite count is $\max(\mathrm{round}(0.1\cdot\text{candidates}),2)$.

\paragraph{Paired evaluation.} For CEM and iCEM, each of six evaluation seeds defines a set of 100 episodes sampled without replacement, and every training configuration replays the same 600 episodes for each solver and compute tier. Planner randomness is initialized from a hash of the episode identifier and tier that is shared across training configurations. All reported planning results pool two independently trained runs per method.

\subsection{Correlation-weight sensitivity}
\label{app:corr-weight}

\begingroup
\newcommand{\abcell}[6]{\shortstack{#1\\[-1pt]\tiny #2/#3/#4/#5/#6}}
\newcommand{\mainabcell}[6]{\shortstack{\textbf{#1}\\[-1pt]\tiny #2/#3/#4/#5/#6}}

\begin{table}[H]
  \centering
  \caption{iCEM sensitivity to the SCALE correlation weight. Each cell reports the success rate (\%) averaged across T1--T5 on the first line, followed by the five tier values. $\dagger$ marks the value used in the main experiments; its averages are bold. We use a single value of $0.1$ across all tasks for a controlled comparison. Selecting $\lambda_{\mathrm{corr}}$ separately for each task gives higher averages on four of five tasks. Results are from a single training run.}
  \label{tab:corr-weight-ablation}
  \scriptsize
  \setlength{\tabcolsep}{2.5pt}
  \renewcommand{\arraystretch}{1.08}
  \resizebox{\linewidth}{!}{%
  \begin{tabular}{lccccc}
    \toprule
    Task & $\lambda_{\mathrm{corr}}{=}0$ & $0.03$ & $0.1^{\dagger}$ & $0.3$ & $1.0$ \\
    \midrule
    Push-T
      & \abcell{63.9}{86}{81}{74}{54}{26} & \abcell{65.8}{88}{83}{76}{55}{26} & \mainabcell{68.2}{90}{87}{79}{60}{26} & \abcell{67.4}{89}{85}{78}{58}{26} & \abcell{68.3}{91}{86}{80}{58}{26} \\
    \midrule
    Reacher
      & \abcell{78.0}{86}{82}{83}{78}{62} & \abcell{79.6}{84}{85}{85}{81}{64} & \mainabcell{81.1}{87}{86}{87}{82}{64} & \abcell{77.3}{82}{84}{81}{78}{62} & \abcell{79.4}{84}{84}{86}{80}{64} \\
    \midrule
    Cube
      & \abcell{65.3}{71}{68}{70}{63}{55} & \abcell{67.8}{73}{74}{72}{64}{57} & \mainabcell{70.5}{77}{76}{75}{65}{60} & \abcell{70.8}{77}{77}{75}{65}{60} & \abcell{70.1}{75}{76}{76}{65}{59} \\
    \midrule
    Two-Room
      & \abcell{82.9}{91}{93}{89}{78}{63} & \abcell{91.7}{100}{100}{96}{88}{76} & \mainabcell{92.9}{100}{100}{97}{90}{77} & \abcell{94.3}{100}{100}{98}{92}{81} & \abcell{95.5}{100}{100}{98}{95}{84} \\
    \midrule
    PointMaze
      & \abcell{80.2}{82}{83}{81}{82}{74} & \abcell{91.7}{97}{96}{94}{88}{83} & \mainabcell{83.4}{87}{83}{84}{83}{79} & \abcell{80.2}{84}{83}{78}{80}{76} & \abcell{75.2}{79}{76}{75}{76}{71} \\
    \bottomrule
  \end{tabular}%
  }
  \normalsize
  \setlength{\abovecaptionskip}{7pt}
  \setlength{\belowcaptionskip}{0pt}
\end{table}
\endgroup

\subsection{Training and planning cost}
\label{app:compute-cost}

Table~\ref{tab:compute-cost} reports training throughput and planning latency on Cube. Planning uses the largest compute tier T1 for every model and solver, corresponding to $45{,}000$ world-model forward passes per plan. At this budget, LeWM amortizes to approximately $27\,\mu$s per world-model forward pass, compared with approximately $2{,}030\,\mu$s for DINO-WM, a roughly $75\times$ difference. 

Training throughput is measured over 300 steps on the same A100 GPU. SCALE adds approximately $13\%$ time per training iteration and Aux adds approximately $4\%$. DINO-WM trains faster because its encoder is frozen.

\begin{table}[H]
  \centering
  \caption{Training throughput and median per-plan latency on Cube. Planning uses the largest matched compute tier T1.}
  \label{tab:compute-cost}
  \scriptsize
  \setlength{\tabcolsep}{3.5pt}
  \renewcommand{\arraystretch}{1.05}
  \resizebox{\linewidth}{!}{%
  \begin{tabular}{lcccc}
    \toprule
    Model & Training (it/s) & CEM (ms/plan) & iCEM (ms/plan) & Planning vs. LeWM \\
    \midrule
    LeWM & $4.35$ & $1{,}213.3$ & $1{,}173.0$ & $1.00\times$ \\
    SCALE & $3.85$ & $1{,}195.8$ & $1{,}184.9$ & $1.00\times$ \\
    Aux & $4.20$ & $1{,}229.4$ & $1{,}218.8$ & $1.03\times$ \\
    DINO-WM & $8.95$ & $91{,}290.7$ & $91{,}654.3$ & $76.7\times$ \\
    \bottomrule
  \end{tabular}%
  }
  \normalsize
  \setlength{\abovecaptionskip}{7pt}
  \setlength{\belowcaptionskip}{0pt}
\end{table}

\FloatBarrier
\section{Additional representation and state-selection results}
\label{app:representation}

Tables~\ref{tab:probing-pusht}--\ref{tab:probing-pointmaze} report held-out $R^2$ for probes from the full frozen embedding to the components of the selected task state $\qvec$ used for supervision. Linear denotes closed-form ridge regression. The MLP is a two-layer network with hidden width 256 and ReLU, trained for 30 epochs using Adam with learning rate $10^{-3}$, batch size 1024, and seed 0. \textcolor{black}{Probe inputs contain all retained principal-component coordinates, divided by one global scalar: 1499 for DINO-WM and approximately 192 for the LeWM-family models. Here full-embedding probing means using all retained components.} The finite-optimization MLP generally has lower absolute $R^2$ than the closed-form probe, so comparisons are made within each probe family. Aux denotes the latent-to-state regression control of Sec.~\ref{sec:method}.

\begin{table}[!htbp]
  \centering
  \caption{Per-component held-out $R^2$ for Push-T. \textcolor{black}{$\vartheta$ is the block angle.}}
  \label{tab:probing-pusht}
  \small
  \begin{tabular}{llcccc}
    \toprule
    Probe & State component & LeWM & SCALE & Aux & DINO-WM \\
    \midrule
    \multirow{7}{*}{Linear}
      & agent $x$       & .9191 & .9656 & \textbf{.9851} & .9342 \\
      & agent $y$       & .9194 & .9715 & \textbf{.9845} & .9394 \\
      & block $x$       & .9686 & .9820 & \textbf{.9904} & .9291 \\
      & block $y$       & .9603 & .9746 & \textbf{.9886} & .9722 \\
      & $\textcolor{black}{\cos\vartheta}$& .9019 & .9023 & \textbf{.9767} & .8353 \\
      & $\textcolor{black}{\sin\vartheta}$& .8694 & .8855 & \textbf{.9778} & .5943 \\
      & \textbf{Mean}   & .9231 & .9469 & \textbf{.9839} & .8674 \\
    \midrule
    \multirow{7}{*}{MLP}
      & agent $x$       & .8580 & .8876 & \textbf{.9187} & .7704 \\
      & agent $y$       & .8676 & \textbf{.9206} & .9066 & .7676 \\
      & block $x$       & .9219 & .9005 & \textbf{.9429} & .7847 \\
      & block $y$       & .9149 & .8943 & \textbf{.9475} & .7685 \\
      & $\textcolor{black}{\cos\vartheta}$& .8572 & .8569 & \textbf{.8980} & .5792 \\
      & $\textcolor{black}{\sin\vartheta}$& .8404 & .8605 & \textbf{.9116} & .4869 \\
      & \textbf{Mean}   & .8767 & .8867 & \textbf{.9209} & .6929 \\
    \bottomrule
  \end{tabular}
  \normalsize
  \setlength{\abovecaptionskip}{7pt}
  \setlength{\belowcaptionskip}{0pt}
\end{table}

\begin{table}[!htbp]
  \centering
  \caption{Per-component held-out $R^2$ for Reacher. $\vartheta_0$ and $\vartheta_1$ denote the two joint angles; the four displayed sine and cosine components constitute the selected joint-only state $\qvec$.}
  \label{tab:probing-reacher}
  \small
  \begin{tabular}{llcccc}
    \toprule
    Probe & State component & LeWM & SCALE & Aux & DINO-WM \\
    \midrule
    \multirow{5}{*}{Linear}
      & $\textcolor{black}{\cos\vartheta_0}$& .9996 & \textbf{.9998} & .9996 & .9972 \\
      & $\textcolor{black}{\sin\vartheta_0}$& .9997 & \textbf{.9998} & \textbf{.9998} & .9981 \\
      & $\textcolor{black}{\cos\vartheta_1}$& .9994 & \textbf{.9997} & .9995 & .9953 \\
      & $\textcolor{black}{\sin\vartheta_1}$& .9980 & \textbf{.9996} & .9986 & .9702 \\
      & \textbf{Mean}   & .9992 & \textbf{.9997} & .9994 & .9902 \\
    \midrule
    \multirow{5}{*}{MLP}
      & $\textcolor{black}{\cos\vartheta_0}$& .9837 & \textbf{.9875} & .9854 & .9479 \\
      & $\textcolor{black}{\sin\vartheta_0}$& .9835 & \textbf{.9882} & .9846 & .9245 \\
      & $\textcolor{black}{\cos\vartheta_1}$& .9780 & .9800 & \textbf{.9820} & .8612 \\
      & $\textcolor{black}{\sin\vartheta_1}$& .9838 & \textbf{.9886} & .9830 & .7814 \\
      & \textbf{Mean}   & .9823 & \textbf{.9861} & .9838 & .8788 \\
    \bottomrule
  \end{tabular}
  \normalsize
  \setlength{\abovecaptionskip}{7pt}
  \setlength{\belowcaptionskip}{0pt}
\end{table}

\begin{table}[!htbp]
  \centering
  \caption{Per-component held-out $R^2$ for Cube. $\varphi$ denotes the gripper yaw; the doubled-angle encoding accounts for the parallel gripper's $\pi$ symmetry. The nine displayed components constitute the selected state $\qvec$.}
  \label{tab:probing-cube}
  \small
  \begin{tabular}{llcccc}
    \toprule
    Probe & State component & LeWM & SCALE & Aux & DINO-WM \\
    \midrule
    \multirow{10}{*}{Linear}
      & effector $x$   & .9773 & .9854 & \textbf{.9934} & .9814 \\
      & effector $y$   & .9661 & .9778 & .9943 & \textbf{.9980} \\
      & effector $z$   & .9825 & .9899 & \textbf{.9928} & .9818 \\
      & $\textcolor{black}{\cos 2\varphi}$& .7599 & .9888 & \textbf{.9938} & .7627 \\
      & $\textcolor{black}{\sin 2\varphi}$& -.2577 & .9880 & \textbf{.9938} & .4213 \\
      & gripper         & .9072 & \textbf{.9783} & .9742 & .6638 \\
      & block $x$       & .9870 & .9922 & \textbf{.9950} & .9282 \\
      & block $y$       & .9884 & .9792 & \textbf{.9924} & .9804 \\
      & block $z$       & .9869 & .9935 & \textbf{.9937} & .8871 \\
      & \textbf{Mean}  & .8109 & .9859 & \textbf{.9915} & .8450 \\
    \midrule
    \multirow{10}{*}{MLP}
      & effector $x$   & \textbf{.8781} & .8326 & .8630 & .7686 \\
      & effector $y$   & .9038 & .9059 & \textbf{.9200} & .8992 \\
      & effector $z$   & .8420 & \textbf{.8603} & .8111 & .8106 \\
      & $\textcolor{black}{\cos 2\varphi}$& .6599 & .9034 & \textbf{.9091} & .5339 \\
      & $\textcolor{black}{\sin 2\varphi}$& -.0394 & \textbf{.8868} & .8724 & .2822 \\
      & gripper         & .7974 & \textbf{.8871} & .8627 & .4786 \\
      & block $x$       & .9083 & .8537 & \textbf{.9145} & .7291 \\
      & block $y$       & \textbf{.9324} & .9081 & .9123 & .8928 \\
      & block $z$       & .8846 & .9024 & \textbf{.9086} & .5871 \\
      & \textbf{Mean}  & .7519 & .8823 & \textbf{.8860} & .6647 \\
    \bottomrule
  \end{tabular}
  \normalsize
  \setlength{\abovecaptionskip}{7pt}
  \setlength{\belowcaptionskip}{0pt}
\end{table}

\begin{table}[!htbp]
  \centering
  \caption{Per-component held-out $R^2$ for Two-Room.}
  \label{tab:probing-tworoom}
  \small
  \begin{tabular}{llcccc}
    \toprule
    Probe & State component & LeWM & SCALE & Aux & DINO-WM \\
    \midrule
    \multirow{3}{*}{Linear}
      & $x$             & .9957 & \textbf{.9982} & .9963 & .9957 \\
      & $y$             & .9942 & .9977 & \textbf{.9981} & .9975 \\
      & \textbf{Mean}  & .9950 & \textbf{.9979} & .9972 & .9966 \\
    \midrule
    \multirow{3}{*}{MLP}
      & $x$             & .9581 & \textbf{.9742} & .9722 & .9315 \\
      & $y$             & .9660 & \textbf{.9829} & .9714 & .9209 \\
      & \textbf{Mean}  & .9621 & \textbf{.9786} & .9718 & .9262 \\
    \bottomrule
  \end{tabular}
  \normalsize
  \setlength{\abovecaptionskip}{7pt}
  \setlength{\belowcaptionskip}{0pt}
\end{table}

\begin{table}[!htbp]
  \centering
  \caption{Per-component held-out $R^2$ for PointMaze.}
  \label{tab:probing-pointmaze}
  \small
  \begin{tabular}{llcccc}
    \toprule
    Probe & State component & LeWM & SCALE & Aux & DINO-WM \\
    \midrule
    \multirow{3}{*}{Linear}
      & $x$             & .9988 & \textbf{.9992} & .9989 & .9982 \\
      & $y$             & .9989 & \textbf{.9990} & .9989 & .9982 \\
      & \textbf{Mean}  & .9989 & \textbf{.9991} & .9989 & .9982 \\
    \midrule
    \multirow{3}{*}{MLP}
      & $x$             & .9703 & \textbf{.9869} & .9717 & .9288 \\
      & $y$             & \textbf{.9824} & .9733 & .9796 & .9609 \\
      & \textbf{Mean}  & .9764 & \textbf{.9801} & .9757 & .9449 \\
    \bottomrule
  \end{tabular}
  \normalsize
  \setlength{\abovecaptionskip}{7pt}
  \setlength{\belowcaptionskip}{0pt}
\end{table}

\subsection{Task-state selection sensitivity}
\label{app:q-selection}

The selected state is defined as follows. For Push-T, $\qvec=(x_{\mathrm{agent}},y_{\mathrm{agent}},x_{\mathrm{block}},y_{\mathrm{block}},\cos\theta_{\mathrm{block}},\sin\theta_{\mathrm{block}})$. For Reacher, $\qvec=(\cos\theta_0,\sin\theta_0,\cos\theta_1,\sin\theta_1)$. For Cube, $\qvec=(x_e,y_e,z_e,\cos 2\psi,\sin 2\psi,g,x_b,y_b,z_b)$ contains the end-effector position, doubled-angle encoding of the gripper yaw, gripper opening, and block position. For Two-Room and PointMaze, $\qvec$ is the two-dimensional agent and ball position, respectively.

The paper selection uses a compact set of static variables, removing velocities and highly redundant coordinates according to a simple task prior. Table~\ref{tab:q-selection-ablation} compares this selection against the full native simulator state on the four tasks where they differ. Two-Room is omitted because its available state has no velocity columns and the paper selection is already the full native state.

\begingroup
\newcommand{\qcell}[3]{\shortstack{$#1{\scriptstyle\pm}#2$\\[-1pt]\tiny $#3$}}
\begin{table}[!htbp]
  \centering
  \caption{Sensitivity to the task-state choice used by SCALE. Each CEM and iCEM cell reports the success rate (\%) averaged across T1--T5 on the first line, followed by the five tier values (mean $\pm$ SD). Results are from a single training run.}
  \label{tab:q-selection-ablation}
  \scriptsize
  \setlength{\tabcolsep}{2.5pt}
  \renewcommand{\arraystretch}{1.12}
  \resizebox{\linewidth}{!}{%
  \begin{tabular}{llccc}
    \toprule
    Task & SCALE state $\qvec$ & $d_q$ & CEM & iCEM \\
    \midrule
    \multirow{2}{*}{Push-T}
      & Selected subset & 6
      & \qcell{73.9}{2.2}{96{\scriptstyle\pm}2.0/93{\scriptstyle\pm}2.6/85{\scriptstyle\pm}4.0/63{\scriptstyle\pm}4.6/32{\scriptstyle\pm}4.4}
      & \qcell{68.2}{2.3}{90{\scriptstyle\pm}2.0/87{\scriptstyle\pm}2.5/79{\scriptstyle\pm}3.9/60{\scriptstyle\pm}4.7/26{\scriptstyle\pm}5.8} \\
      & Full state & 8
      & \qcell{74.5}{2.6}{94{\scriptstyle\pm}2.9/92{\scriptstyle\pm}2.6/85{\scriptstyle\pm}3.0/67{\scriptstyle\pm}5.0/34{\scriptstyle\pm}5.8}
      & \qcell{68.3}{2.8}{90{\scriptstyle\pm}1.6/87{\scriptstyle\pm}1.8/79{\scriptstyle\pm}4.8/59{\scriptstyle\pm}6.9/26{\scriptstyle\pm}1.2} \\
    \midrule
    \multirow{2}{*}{Reacher}
      & Selected subset & 4
      & \qcell{71.5}{3.9}{82{\scriptstyle\pm}4.6/79{\scriptstyle\pm}3.8/75{\scriptstyle\pm}5.9/64{\scriptstyle\pm}8.4/57{\scriptstyle\pm}6.1}
      & \qcell{81.1}{2.0}{87{\scriptstyle\pm}3.6/86{\scriptstyle\pm}2.9/86{\scriptstyle\pm}3.4/82{\scriptstyle\pm}3.2/64{\scriptstyle\pm}2.5} \\
      & Full state & 8
      & \qcell{72.1}{1.0}{82{\scriptstyle\pm}2.4/81{\scriptstyle\pm}1.0/73{\scriptstyle\pm}2.8/66{\scriptstyle\pm}3.4/59{\scriptstyle\pm}6.1}
      & \qcell{82.3}{1.4}{89{\scriptstyle\pm}2.3/86{\scriptstyle\pm}2.4/89{\scriptstyle\pm}2.9/84{\scriptstyle\pm}4.9/64{\scriptstyle\pm}2.7} \\
    \midrule
    \multirow{2}{*}{Cube}
      & Selected subset & 9
      & \qcell{64.1}{4.0}{74{\scriptstyle\pm}3.4/71{\scriptstyle\pm}4.2/66{\scriptstyle\pm}4.8/56{\scriptstyle\pm}5.2/54{\scriptstyle\pm}5.0}
      & \qcell{70.5}{3.3}{77{\scriptstyle\pm}3.9/76{\scriptstyle\pm}4.5/75{\scriptstyle\pm}3.5/65{\scriptstyle\pm}3.7/60{\scriptstyle\pm}4.5} \\
      & Full state & 22
      & \qcell{64.8}{3.1}{76{\scriptstyle\pm}5.0/71{\scriptstyle\pm}3.5/64{\scriptstyle\pm}3.9/58{\scriptstyle\pm}4.1/55{\scriptstyle\pm}4.8}
      & \qcell{69.7}{3.4}{76{\scriptstyle\pm}3.7/76{\scriptstyle\pm}4.2/74{\scriptstyle\pm}4.1/64{\scriptstyle\pm}4.5/57{\scriptstyle\pm}3.5} \\
    \midrule
    \multirow{2}{*}{PointMaze}
      & Selected subset & 2
      & \qcell{86.2}{2.4}{89{\scriptstyle\pm}3.7/87{\scriptstyle\pm}5.1/88{\scriptstyle\pm}3.3/86{\scriptstyle\pm}2.3/80{\scriptstyle\pm}1.9}
      & \qcell{83.4}{2.2}{87{\scriptstyle\pm}3.9/83{\scriptstyle\pm}3.9/84{\scriptstyle\pm}4.1/83{\scriptstyle\pm}4.6/79{\scriptstyle\pm}2.4} \\
      & Full state & 4
      & \qcell{86.2}{3.6}{88{\scriptstyle\pm}4.9/88{\scriptstyle\pm}4.9/89{\scriptstyle\pm}4.4/84{\scriptstyle\pm}4.4/82{\scriptstyle\pm}4.5}
      & \qcell{86.0}{4.1}{88{\scriptstyle\pm}4.5/88{\scriptstyle\pm}3.7/87{\scriptstyle\pm}4.2/85{\scriptstyle\pm}3.9/81{\scriptstyle\pm}6.8} \\
    \bottomrule
  \end{tabular}%
  }
  \normalsize
  \setlength{\abovecaptionskip}{7pt}
  \setlength{\belowcaptionskip}{0pt}
\end{table}
\endgroup

The paper and full-state variants are close relative to the reported variability. Seven of the eight task--solver comparisons differ by at most $1.2$ percentage points in overall success; the largest difference is the $+2.6$-point change for PointMaze with iCEM, while Cube with iCEM changes by $-0.8$ points. There is therefore no consistent degradation from including the native velocity or redundant configuration variables in the distance target. These results indicate that SCALE is not sensitive to the particular compact state selection used in the paper.

\subsection{Held-out latent--state distance alignment}
\label{app:distance-alignment}

\textcolor{black}{We assess held-out distance alignment using Spearman rank correlation $\rho_{\mathrm{S}}$ under the protocol of Sec.~\ref{sec:geometry}. Whereas SCALE's loss uses Pearson correlation $\rho_{\mathrm{P}}$ of distance magnitudes, this diagnostic measures agreement between their ranks. It is therefore a complementary check of monotonic alignment, not the exact quantity optimized by $\Lcorr$, and does not by itself establish the mechanism examined in Sec.~\ref{sec:validation}.}

\begin{table}[!htbp]
  \centering
  \caption{Held-out latent--state distance rank alignment (Spearman~$\textcolor{black}{\rho_{\mathrm{S}}}$) over $1.12$M frame pairs per task. LeWM and DINO-WM repeat Table~\ref{tab:ref-align} for comparison.}
  \label{tab:heldout-corr}
  \small
  \begin{tabular}{lcccc}
    \toprule
    Task & LeWM & SCALE & Aux & DINO-WM \\
    \midrule
    Push-T    & .13 & \textbf{.77} & .18 & .70 \\
    Reacher   & .18 & \textbf{.53} & .17 & .46 \\
    Cube      & .00 & \textbf{.56} & .03 & .46 \\
    Two-Room  & .41 & .77 & .41 & \textbf{.85} \\
    PointMaze & .52 & \textbf{.80} & .52 & .50 \\
    \bottomrule
  \end{tabular}
  \normalsize
  \setlength{\abovecaptionskip}{7pt}
  \setlength{\belowcaptionskip}{0pt}
\end{table}

SCALE substantially increases latent--state rank alignment over LeWM on every task. Aux, in contrast, leaves the alignment essentially unchanged despite making task state highly recoverable from the full embedding. The latter provides an additional check on the distinction motivating SCALE: state decodability alone does not ensure that state differences are expressed in the latent metric.

\FloatBarrier
\section{Full planning results}
\label{app:full-results}

Tables~\ref{tab:full-pusht}--\ref{tab:full-pointmaze} report success rates for every task, planning solver, and compute tier, including DINO-WM as the pretrained-feature reference. Each tier pools two training runs evaluated on six paired sets, giving $n=12$ 100-episode measurements. The Overall column averages T1--T5 within each replicate.

\begin{table}[!htbp]
  \centering
  \caption{Pooled Push-T success rates (\%). Mean $\pm$ SD over two training runs and six evaluation sets per run. The Overall column pools T1--T5. Bold marks the best method per column within each solver.}
  \label{tab:full-pusht}
  \scriptsize
  \setlength{\tabcolsep}{2.5pt}
  \renewcommand{\arraystretch}{0.92}
  \begin{tabular}{llcccccc}
    \toprule
    Solver & Method & T1 & T2 & T3 & T4 & T5 & Overall \\
    \midrule
    \multirow{4}{*}{CEM}
      & LeWM & \sr{92.3}{2.3} & \sr{89.3}{3.2} & \sr{80.5}{4.1} & \sr{57.8}{5.6} & \sr{33.8}{3.6} & \sr{70.7}{2.7} \\
      & SCALE & \bsr{94.8}{3.2} & \sr{92.1}{2.6} & \sr{84.2}{3.6} & \sr{61.6}{6.9} & \sr{32.8}{5.5} & \sr{73.1}{2.7} \\
      & Aux & \sr{94.5}{2.0} & \bsr{92.6}{2.4} & \bsr{84.4}{4.1} & \bsr{65.1}{6.9} & \bsr{34.9}{6.4} & \bsr{74.3}{3.4} \\
      & DINO-WM & \sr{74.8}{4.6} & \sr{72.1}{4.5} & \sr{65.2}{4.9} & \sr{46.1}{4.6} & \sr{24.3}{6.1} & \sr{56.5}{4.0} \\
    \midrule
    \multirow{4}{*}{iCEM}
      & LeWM & \sr{86.0}{2.4} & \sr{79.8}{2.1} & \sr{73.4}{5.9} & \sr{55.2}{4.8} & \sr{25.1}{5.1} & \sr{63.9}{3.0} \\
      & SCALE & \sr{89.1}{2.0} & \bsr{86.3}{2.8} & \bsr{79.7}{3.6} & \sr{61.3}{6.2} & \sr{25.2}{5.5} & \sr{68.3}{2.8} \\
      & Aux & \bsr{89.9}{2.0} & \sr{86.2}{2.3} & \sr{79.6}{3.4} & \bsr{62.2}{5.3} & \bsr{27.6}{5.2} & \bsr{69.1}{2.6} \\
      & DINO-WM & \sr{70.0}{4.5} & \sr{65.3}{4.5} & \sr{58.8}{5.1} & \sr{41.5}{8.8} & \sr{23.1}{3.7} & \sr{51.7}{4.3} \\
    \bottomrule
  \end{tabular}
  \normalsize
  \setlength{\abovecaptionskip}{7pt}
  \setlength{\belowcaptionskip}{0pt}
\end{table}

\begin{table}[!htbp]
  \centering
  \caption{Pooled Reacher success rates (\%). Mean $\pm$ SD over two training runs and six evaluation sets per run. The Overall column pools T1--T5. Bold marks the best method per column within each solver.}
  \label{tab:full-reacher}
  \scriptsize
  \setlength{\tabcolsep}{2.5pt}
  \renewcommand{\arraystretch}{0.92}
  \begin{tabular}{llcccccc}
    \toprule
    Solver & Method & T1 & T2 & T3 & T4 & T5 & Overall \\
    \midrule
    \multirow{4}{*}{CEM}
      & LeWM & \bsr{82.3}{1.8} & \bsr{79.4}{3.6} & \sr{73.3}{4.8} & \bsr{66.2}{4.1} & \sr{55.8}{5.6} & \sr{71.4}{2.3} \\
      & SCALE & \sr{82.2}{3.8} & \sr{79.1}{5.1} & \sr{74.5}{4.4} & \sr{65.2}{6.0} & \bsr{57.4}{5.5} & \bsr{71.7}{3.1} \\
      & Aux & \sr{81.6}{4.6} & \sr{78.8}{4.2} & \bsr{74.9}{3.2} & \sr{65.8}{5.6} & \sr{55.9}{6.8} & \sr{71.4}{3.3} \\
      & DINO-WM & \sr{77.2}{2.8} & \sr{74.8}{5.4} & \sr{70.8}{5.9} & \sr{61.8}{3.4} & \sr{50.2}{7.3} & \sr{67.0}{3.0} \\
    \midrule
    \multirow{4}{*}{iCEM}
      & LeWM & \sr{84.9}{2.6} & \sr{84.2}{3.4} & \sr{83.8}{3.4} & \sr{79.0}{6.1} & \sr{62.2}{3.6} & \sr{78.8}{1.7} \\
      & SCALE & \bsr{86.7}{3.6} & \bsr{87.0}{3.1} & \bsr{87.3}{3.2} & \bsr{81.7}{3.0} & \bsr{64.4}{3.0} & \bsr{81.4}{2.0} \\
      & Aux & \sr{85.0}{4.4} & \sr{86.1}{3.1} & \sr{85.8}{3.5} & \sr{80.1}{3.7} & \sr{62.9}{5.1} & \sr{80.0}{2.2} \\
      & DINO-WM & \sr{78.7}{4.9} & \sr{78.2}{4.8} & \sr{80.8}{3.5} & \sr{76.0}{4.0} & \sr{53.0}{5.5} & \sr{73.3}{3.1} \\
    \bottomrule
  \end{tabular}
  \normalsize
  \setlength{\abovecaptionskip}{7pt}
  \setlength{\belowcaptionskip}{0pt}
\end{table}

\begin{table}[!htbp]
  \centering
  \caption{Pooled Cube success rates (\%). Mean $\pm$ SD over two training runs and six evaluation sets per run. The Overall column pools T1--T5. Bold marks the best method per column within each solver.}
  \label{tab:full-cube}
  \scriptsize
  \setlength{\tabcolsep}{2.5pt}
  \renewcommand{\arraystretch}{0.92}
  \begin{tabular}{llcccccc}
    \toprule
    Solver & Method & T1 & T2 & T3 & T4 & T5 & Overall \\
    \midrule
    \multirow{4}{*}{CEM}
      & LeWM & \sr{67.6}{4.6} & \sr{65.8}{3.8} & \sr{60.8}{4.6} & \bsr{56.8}{3.5} & \sr{53.8}{4.0} & \sr{60.9}{3.4} \\
      & SCALE & \bsr{73.9}{3.9} & \bsr{71.1}{4.1} & \bsr{64.7}{3.8} & \sr{55.9}{4.7} & \sr{54.8}{4.6} & \bsr{64.1}{3.6} \\
      & Aux & \sr{71.6}{4.7} & \sr{70.2}{4.5} & \sr{62.8}{5.2} & \sr{56.6}{4.6} & \bsr{55.2}{4.3} & \sr{63.3}{3.6} \\
      & DINO-WM & \sr{73.4}{3.1} & \sr{66.9}{2.3} & \sr{58.4}{4.4} & \sr{51.6}{4.5} & \sr{53.9}{3.8} & \sr{60.9}{2.9} \\
    \midrule
    \multirow{4}{*}{iCEM}
      & LeWM & \sr{68.9}{4.9} & \sr{67.1}{3.6} & \sr{68.7}{4.1} & \sr{61.9}{3.9} & \sr{54.9}{5.7} & \sr{64.3}{4.0} \\
      & SCALE & \bsr{76.8}{3.7} & \bsr{76.6}{4.0} & \bsr{75.2}{3.0} & \bsr{65.8}{3.3} & \bsr{59.8}{3.9} & \bsr{70.8}{3.1} \\
      & Aux & \sr{73.8}{3.6} & \sr{73.0}{4.7} & \sr{72.5}{3.8} & \sr{64.8}{3.4} & \sr{57.8}{3.8} & \sr{68.4}{3.5} \\
      & DINO-WM & \sr{69.9}{2.5} & \sr{67.4}{1.6} & \sr{66.0}{1.7} & \sr{57.7}{3.8} & \sr{51.4}{3.6} & \sr{62.5}{1.3} \\
    \bottomrule
  \end{tabular}
  \normalsize
  \setlength{\abovecaptionskip}{7pt}
  \setlength{\belowcaptionskip}{0pt}
\end{table}

\begin{table}[!htbp]
  \centering
  \caption{Pooled Two-Room success rates (\%). Mean $\pm$ SD over two training runs and six evaluation sets per run. The Overall column pools T1--T5. Bold marks the best method per column within each solver.}
  \label{tab:full-tworoom}
  \scriptsize
  \setlength{\tabcolsep}{2.5pt}
  \renewcommand{\arraystretch}{0.92}
  \begin{tabular}{llcccccc}
    \toprule
    Solver & Method & T1 & T2 & T3 & T4 & T5 & Overall \\
    \midrule
    \multirow{4}{*}{CEM}
      & LeWM & \sr{87.8}{3.0} & \sr{87.1}{2.8} & \sr{83.0}{3.3} & \sr{71.8}{2.7} & \sr{57.8}{3.9} & \sr{77.5}{1.5} \\
      & SCALE & \sr{99.7}{0.7} & \sr{99.3}{0.8} & \sr{94.0}{2.0} & \sr{83.5}{2.6} & \sr{69.7}{4.2} & \sr{89.2}{0.8} \\
      & Aux & \sr{88.3}{2.7} & \sr{87.7}{2.9} & \sr{82.4}{3.1} & \sr{70.6}{3.1} & \sr{59.7}{2.5} & \sr{77.7}{1.3} \\
      & DINO-WM & \bsr{100.0}{0.0} & \bsr{99.8}{0.4} & \bsr{98.2}{1.2} & \bsr{94.1}{2.2} & \bsr{85.8}{3.7} & \bsr{95.6}{1.0} \\
    \midrule
    \multirow{4}{*}{iCEM}
      & LeWM & \sr{91.1}{1.6} & \sr{92.9}{1.7} & \sr{88.5}{2.0} & \sr{77.3}{2.2} & \sr{62.4}{2.8} & \sr{82.5}{1.2} \\
      & SCALE & \sr{99.9}{0.3} & \sr{99.8}{0.6} & \sr{97.7}{1.2} & \sr{90.2}{3.0} & \sr{76.2}{2.7} & \sr{92.7}{1.0} \\
      & Aux & \sr{92.3}{2.1} & \sr{93.2}{1.7} & \sr{89.2}{2.0} & \sr{78.7}{2.5} & \sr{62.9}{3.6} & \sr{83.3}{1.1} \\
      & DINO-WM & \bsr{100.0}{0.0} & \bsr{100.0}{0.0} & \bsr{99.8}{0.5} & \bsr{98.6}{0.8} & \bsr{92.5}{2.1} & \bsr{98.2}{0.5} \\
    \bottomrule
  \end{tabular}
  \normalsize
  \setlength{\abovecaptionskip}{7pt}
  \setlength{\belowcaptionskip}{0pt}
\end{table}

\begin{table}[!htbp]
  \centering
  \caption{Pooled PointMaze success rates (\%). Mean $\pm$ SD over two training runs and six evaluation sets per run. The Overall column pools T1--T5. Bold marks the best method per column within each solver.}
  \label{tab:full-pointmaze}
  \scriptsize
  \setlength{\tabcolsep}{2.5pt}
  \renewcommand{\arraystretch}{0.92}
  \begin{tabular}{llcccccc}
    \toprule
    Solver & Method & T1 & T2 & T3 & T4 & T5 & Overall \\
    \midrule
    \multirow{4}{*}{CEM}
      & LeWM & \sr{85.7}{4.9} & \sr{86.0}{4.2} & \sr{86.9}{3.9} & \sr{83.5}{3.5} & \sr{75.3}{5.8} & \sr{83.5}{3.3} \\
      & SCALE & \sr{87.6}{3.8} & \sr{87.4}{3.9} & \sr{86.8}{3.8} & \sr{84.8}{3.9} & \sr{78.1}{3.4} & \sr{84.9}{2.7} \\
      & Aux & \sr{85.0}{4.6} & \sr{83.9}{4.7} & \sr{82.7}{4.8} & \sr{79.8}{4.7} & \sr{78.1}{5.6} & \sr{81.9}{3.9} \\
      & DINO-WM & \bsr{98.2}{1.2} & \bsr{98.3}{1.4} & \bsr{97.2}{1.3} & \bsr{92.4}{3.0} & \bsr{87.2}{2.9} & \bsr{94.7}{1.4} \\
    \midrule
    \multirow{4}{*}{iCEM}
      & LeWM & \sr{84.0}{4.1} & \sr{83.6}{4.3} & \sr{82.4}{3.1} & \sr{81.2}{4.6} & \sr{73.2}{4.8} & \sr{80.9}{3.5} \\
      & SCALE & \sr{87.6}{3.3} & \sr{85.8}{4.5} & \sr{84.6}{3.2} & \sr{81.9}{4.7} & \sr{77.0}{4.4} & \sr{83.4}{2.5} \\
      & Aux & \sr{80.8}{5.1} & \sr{79.5}{5.1} & \sr{78.7}{5.0} & \sr{79.4}{3.7} & \sr{75.0}{5.8} & \sr{78.7}{4.3} \\
      & DINO-WM & \bsr{93.4}{2.6} & \bsr{92.2}{2.6} & \bsr{90.8}{3.4} & \bsr{90.3}{2.6} & \bsr{85.2}{2.6} & \bsr{90.4}{2.1} \\
    \bottomrule
  \end{tabular}
  \normalsize
  \setlength{\abovecaptionskip}{7pt}
  \setlength{\belowcaptionskip}{0pt}
\end{table}

\end{document}